\documentclass[11pt]{article}
\usepackage[table]{xcolor}

\usepackage[final]{acl}

\usepackage{times}
\usepackage{latexsym}
\usepackage{booktabs}
\usepackage{amsmath}
\usepackage{amssymb}
\usepackage{adjustbox}
\usepackage{multirow}
\usepackage{hyperref}
\usepackage{titletoc} 
\usepackage{etoc}

\usepackage{tabularx}
\newcolumntype{Y}{>{\centering\arraybackslash}X}

\usepackage{tikz}
\usepackage{amsmath,amssymb}
\usepackage{xcolor}
\usetikzlibrary{shapes.geometric, arrows.meta, positioning, calc, fit,
                backgrounds, decorations.pathreplacing}

\definecolor{nircBlue}{RGB}{170, 195, 230}
\definecolor{nircBlueD}{RGB}{60, 95, 160}
\definecolor{nircGreen}{RGB}{175, 220, 175}
\definecolor{nircGreenD}{RGB}{50, 130, 70}
\definecolor{nircGray}{RGB}{235, 235, 235}
\definecolor{nircGrayD}{RGB}{90, 90, 100}
\definecolor{nircOrange}{RGB}{250, 200, 150}
\definecolor{nircOrangeD}{RGB}{200, 110, 40}

\usepackage{xspace}
\usepackage{comment}
\definecolor{lightgrey}{rgb}{0.9, 0.9, 0.9}
\definecolor{lightgreen}{rgb}{0.85, 0.95, 0.85}

\newcommand{\methodname}{\textsc{SubFit}\xspace}
\newcommand{\methodnamefull}{Submodule-level Fitted residual replacement\xspace}
\newcommand{\std}[1]{{\tiny$\pm$#1}}

\usepackage[T1]{fontenc}

\usepackage[utf8]{inputenc}

\usepackage{microtype}

\usepackage{inconsolata}

\usepackage{graphicx}
\usepackage{cleveref}
\usepackage{enumitem}
\usepackage{amsmath}

\title{From Layers to Submodules: Rethinking Granularity in Replacement-Based LLM Compression}

\author{ \quad
Elia Cunegatti \quad
Marcus Vukojevic \quad
Erik Nielsen \quad
Giovanni Iacca \\
University of Trento \\
\texttt{\{elia.cunegatti, marcus.vukojevic, erik.nielsen, giovanni.iacca\}@unitn.it}
}

\begin{document}
\maketitle

\begin{abstract}
Post-training compression of Large Language Models (LLMs) removes entire architectural components, either deleting them or replacing them with fitted modules. Existing replacement-based methods share two design constraints: full-layer granularity and contiguous selection. We argue that this is overly restrictive: in fact, redundancy in pretrained transformers is not confined to contiguous regions, nor does it evenly distribute between Attention and FeedForward outputs, implying that different strategies best approximate different submodule types and that removable components need not cluster within contiguous depth ranges. Based on this intuition, we introduce \methodname (\methodnamefull), which compresses LLMs at the \emph{submodule} level: Attention and FeedForward submodules are selected non-contiguously, and each receives its own lightweight fitted residual bypass. \methodname operates post-training and requires only calibration data. Across ten LLMs (five base, five instruction-tuned), five sparsity levels from $12.5\%$ to $37.5\%$, and four replacement-based baselines, \methodname achieves the best aggregate perplexity–accuracy trade-off across the evaluated sparsity levels, with larger gains under aggressive compression. At $25\%$ sparsity, it retains $84.6\%$ of dense downstream accuracy and incurs $2.42\times$ perplexity degradation, against $81.6\%$ and $4.34\times$ for the strongest baselines, while delivering measurable inference speedup and KV-cache savings. Code is available at \href{https://github.com/eliacunegatti/SubFit}{https://github.com/eliacunegatti/SubFit}.
\end{abstract}

\section{Introduction}

\begin{figure}[!ht]
    \setlength{\belowcaptionskip}{-10pt}
    \centering
    \includegraphics[width=\linewidth]{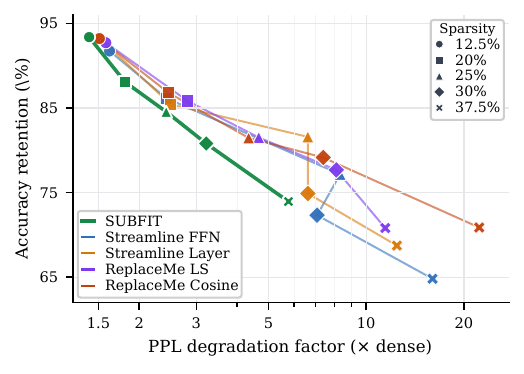}
    \caption{Downstream accuracy retention versus aggregate perplexity degradation across sparsity levels for \methodname and four replacement-based baselines.}
    \label{fig:teaser}
\end{figure}

Post-training compression is an established approach to deploying Large Language Models (LLMs) under tight memory and latency constraints~\cite{lin2024awq,xiao2023smoothquant}. Methods that remove entire architectural components~\cite{ma2023llm,ashkboos2024slicegpt} are particularly effective because they yield real inference savings without depending on sparse kernels or custom hardware. However, the challenge is that removing components from a pretrained model disrupts the residual stream, and na\"ive deletion typically incurs sharp quality loss. 

A recent line of work addresses this by replacing removed components with lightweight fitted modules or recovery mechanisms~\citep{streamline,replaceme,liu2025grasp,chen2026simple,jiang2026equilibrium}. While these approaches differ in whether they use calibration data, fold the replacement into existing weights, or apply activation correction at inference time, they share the same core observation: recovery after component removal is crucial.
However, most operate at the granularity of full layers or contiguous blocks, treating the Transformer layer as the atomic unit of removal. Submodule-level granularity has already been explored in deletion-based structured pruning: BlockPruner~\cite{zhong-etal-2025-blockpruner} and 2SSP~\cite{sandri2025ssp} independently remove Attention and FFN submodules rather than entire Transformer layers, demonstrating the viability of finer-grained selection. 
In this work, we bring this granularity to replacement-based compression: the goal is not only to remove components, but to identify components whose residual contribution can be approximated efficiently. Attentions require only a low-rank bypass, while FeedForwards (FFNs) need a higher-rank map whose input basis is shared across selected layers to limit deployed cost. 

Based on these observations, we introduce \methodname (\methodnamefull), a replacement method that acts at the Attention and FFN submodule-level, selecting components across depths and fitting a residual bypass for each removed submodule. We evaluate the method on five base LLMs, five instruction-tuned variants, and five sparsity levels from $12.5\%$ to $37.5\%$, against four post-training, calibration-only replacement baselines from the LLM-Streamline \cite{streamline} (henceforth referred to simply as ``Streamline'') and ReplaceMe \cite{replaceme} families. At the aggregate level, i.e., averaging perplexity (PPL) degradation over base models and accuracy retention over instruction-tuned models, \methodname achieves the strongest perplexity--accuracy trade-off across the evaluated sparsity levels (\Cref{fig:teaser}), with larger gains under aggressive compression: the PPL gap over the strongest baseline grows from $0.11\times$ at $12.5\%$ to $1.92\times$ at $25\%$ and $5.69\times$ at $37.5\%$, while also obtaining the most stable results. 
The same pattern holds for downstream accuracy retention on instruction-tuned models, where \methodname is the only method to retain above $80\%$ at $25\%$ sparsity and above $73\%$ at $37.5\%$. 

To summarize, our contributions are as follows. 
\textbf{(1)} We revisit submodule-level compression in the replacement-based setting and empirically examine the effect of combining fine-grained granularity with non-contiguous selection. \textbf{(2)} We propose \methodname, which selects replaceable Attention and FFN submodules and fits a residual bypass for each, using a shared low-rank basis across FFNs to limit deployed cost. 
\textbf{(3)} We evaluate \methodname on ten LLMs and five sparsity levels, reporting perplexity, downstream accuracy, deployed cost, and inference speedup; results show improved aggregate quality--efficiency trade-offs under increasing compression. We also analyze the selection criteria, calibration sensitivity, and cost of the proposed method.

\section{Related Work}

\noindent\textbf{Post-training LLM compression.}
Post-training compression reduces the deployment cost of pretrained LLMs without requiring retraining. \emph{Quantization} methods reduce the numerical precision of weights, exploiting second-order information~\citep{frantar2022gptq, kim2024squeezellm} as well as activation statistics~\citep{lin2024awq}. Other approaches jointly compress weights and activations, improving throughput through outlier-aware corrections~\citep{xiao2023smoothquant, shao2024omniquant}. On the other hand, \emph{unstructured pruning} approaches, such as SparseGPT~\citep{frantar2023sparsegpt} and Wanda~\citep{sun2024wanda}, directly remove weights from the model matrices. However, weight removal often requires sparse kernels or hardware support to translate into a practical runtime speedup. For deployable compression, \emph{structured pruning} is often preferred because it removes entire architectural units, such as rows or columns of a weight matrix~\citep{ashkboos2024slicegpt, an2024fluctuation}, or complete components such as Transformer blocks~\citep{yang2024laco,gromov2025the,men2025shortgpt},
Attention heads~\citep{ma2023llm}. Submodule-level removal has been shown to be a promising structured-pruning strategy in BlockPruner~\citep{zhong-etal-2025-blockpruner} and 2SSP~\citep{sandri2025ssp}, where Attention and FFN submodules are removed across depth. However, like the other methods described above, they delete the selected components without accounting for how accurately their residual contributions can be approximated after removal. \methodname adopts the same fine-grained removal units but applies them in the replacement-based setting, selecting submodules according to their replacement behavior and fitting a dedicated residual bypass for each removed component.

\noindent\textbf{Compression during training.}
Another family of compression approaches relies on jointly compressing and training. Depending on the method, this can involve layer sharing~\citep{xia2024sheared}, sparsity-aware training~\citep{liu2025pat,tang2025darwinlm}, or quantization-aware training~\citep{liu2024llm,chen2025efficientqat}. These approaches let the model adapt to the compressed structure during optimization, but require additional training resources and runtime, unlike \methodname.

\noindent\textbf{Block replacement and post-pruning recovery.}
Replacement-based methods are a recent direction in post-training compression, where the key observation is that recovering the contribution of removed components can improve task performance without full-model fine-tuning and without re-introducing the original parameters. These methods remove Transformer blocks or layers and replace them with lightweight alternatives fitted on calibration data: a small feed-forward module~\citep{streamline}, a linear transformation that can be absorbed into adjacent
weights~\citep{replaceme}, SVD~\citep{liu2025grasp}, or a lightweight fixed-point module selected by a learned policy~\citep{jiang2026equilibrium}. LinearPatch~\citep{chen2026simple} instead targets already layer-pruned models by correcting activation-magnitude mismatch through a linear patch between the remaining blocks. These methods show that recovery after component removal is crucial, but they primarily operate at the granularity of full layers or contiguous blocks. Hence, the novelty of \methodname lies not in submodule-level removal alone, but in combining this finer granularity with independent non-contiguous selection, selection criteria tailored to post-replacement behavior, and fitted per-submodule residual compensation.

\section{Methodology}
\label{sec:method}
\methodname compresses pretrained decoder-only Transformer architectures through a two-stage \emph{select-and-replace} procedure. Rather than considering deletion tolerance alone, \methodname treats \emph{replaceability} as the relevant selection objective: a submodule is considered replaceable when its residual contribution can be accurately approximated by the fitted surrogate under the target compression budget. Given a target sparsity $s$ and a model with $L$ layers, \methodname selects $\mathrm{round}(Ls)$ Attention submodules and $\mathrm{round}(Ls)$ FFN submodules independently across depths, and replaces their residual contributions with lightweight fitted bypasses.
Selection is non-contiguous: each selected module is chosen independently by score, with no restriction that selected layers form a consecutive block.

\noindent\textbf{Notation.}
For a submodule $F_\ell$ at layer $\ell$, either Attention or FFN, we denote by $x \in \mathbb{R}^d$ a generic token-level input and by $y = F_\ell(x) \in \mathbb{R}^d$ the corresponding residual contribution to be approximated, where $d$ is the hidden dimension, shared by both input and output of each submodule.
Calibration samples are written $(x_i, y_i)_{i=1}^{N}$ with $y_i = F_\ell(x_i)$, where $N$ is the number of calibration tokens, and we collect them into matrices $X_\ell, Y_\ell \in \mathbb{R}^{N \times d}$ to generate our dataset for the fitting stage.
The sets of selected Attention and FFN layers are denoted by $\mathcal{S}_{\mathrm{attn}}$ and $\mathcal{S}_{\mathrm{fn}}$, respectively.

\noindent\textbf{Sequential component selection.}
\methodname scores and replaces Attention submodules first, then scores FFN submodules on the partially compressed model. This order reflects two properties of pretrained Transformers: Attention submodules tend to be more redundant than FFN submodules~\citep{he2024matterstransformersattentionneeded,siddiqui2024deeper}, and scoring the second component on post-replacement activations lets FFN selection account for the residual-stream shift induced by Attention replacement, as in activation-driven post-training compression~\citep{frantar2023sparsegpt,sun2024wanda,ashkboos2024slicegpt}. For Attentions, we rank submodules by an \emph{Impact} score computed on calibration activations. Given a residual reference $h$ and a module contribution $\delta$, the per-token score is
\begin{equation}
I(h,\delta) = \bigl(1-\cos(h,\,h+\delta)\bigr)\,
\frac{\|\delta\|_2}{\|h\|_2+\epsilon},
\label{eq:impact}
\end{equation}
where $\epsilon\!>\!0$ is a small constant for numerical stability. We use Eq.~\eqref{eq:impact} as a residual-stream perturbation proxy rather than as a theoretically optimal importance criterion. The cosine term captures the directional change induced in the residual stream~\citep{gromov2025the,streamline}, while the norm ratio weights this change by the relative magnitude of the submodule contribution~\citep{he2024matterstransformersattentionneeded}. Their product is small when the submodule induces a limited directional or magnitude perturbation, identifying contributions that can be approximated by lightweight surrogates of the form in Eq.~\eqref{eq:generic_surrogate}.
We select Attention modules with the lowest impact scores, aggregated robustly across calibration tokens. 
On the other hand, FFN selection differs between base and instruction-tuned models.
For base models, we use a \emph{replacement-aware} score that combines the effect of removing the FFN contribution with the residual error of the fitted surrogate, in the spirit of reconstruction-error importance~\citep{frantar2023sparsegpt,sun2024wanda}.
For instruction-tuned models, we find that FFN residual magnitudes after SFT can be poorly calibrated across depths, making norm-based factors noisy~\citep{harada2025massive}; we therefore use a cosine-only score, also used in ReplaceMe~\citep{replaceme}, retaining only the cosine factor of Eq.~\eqref{eq:impact}. The controlled comparison in \Cref{tab:ffn-scoring-comparison} supports this FFN selection policy: replacement-aware scoring yields lower aggregate PPL for base models, whereas cosine scoring achieves higher average accuracy for instruction-tuned models.

\noindent\textbf{Fitted residual replacement.}
For each selected submodule, we fit a parametric surrogate $\widetilde{F}_\ell(\cdot\,;\theta_\ell)$, defined in detail in the next paragraph, by minimizing the residual error on calibration pairs:
\begin{equation}
\theta_\ell^\star = \arg\min_{\theta_\ell}
\sum_{i=1}^{N} \bigl\|
\widetilde{F}_\ell(x_i;\theta_\ell)-y_i \bigr\|_2^2,
\label{eq:generic_fit}
\end{equation}
with $\theta_\ell^\star$ recovered analytically from calibration statistics, without full-model fine-tuning, in the spirit of activation-driven post-training compression~\citep{frantar2023sparsegpt, sun2024wanda}.

\noindent\textbf{A common submodule surrogate.}
Both Attention and FFN submodules contribute to the residual stream via an additive map
$\mathbb{R}^d\!\to\!\mathbb{R}^d$
\citep{vaswani2017attention}; since each submodule $F_\ell$ communicates with the rest of the network only through this output, we approximate it through its input-output signature rather than reproducing its internal computation. For both kinds of submodules, we use the same surrogate structure:
\begin{equation}
\widetilde{F}_\ell(x;\theta_\ell) \;=\;
\underbrace{g_\ell \odot x + b_\ell}_{\text{affine correction}}
\;+\;
\underbrace{(x - \mu_\ell)\,U_\ell}_{\text{structured interaction}},
\label{eq:generic_surrogate}
\end{equation}
where $\theta_\ell = \{g_\ell, b_\ell, \mu_\ell, U_\ell\}$ collects all layer specific parameters, $\odot$ denotes the element-wise (Hadamard) product, and $U_\ell \in \mathbb{R}^{d \times d}$ is a  low-rank operator constrained by $\operatorname{rank}(U_\ell) \le r$, factored as:
\begin{equation}
U_\ell = V_\ell^{\!\top} W_\ell, \qquad
V_\ell, W_\ell \in \mathbb{R}^{r \times d}.
\label{eq:lowrank_factor}
\end{equation}
Each component plays a distinct role. 
The element-wise gate $g_\ell$ restores the per-feature scale of the removed contribution, accounting for the fact that different hidden dimensions are written into the residual stream with different magnitudes. 
The bias $b_\ell$ and the centering term $\mu_\ell$ together absorb the mean shift induced by removing the original submodule output~\citep{xu2025prunecomp}: $\mu_\ell$ re-centers the input around the calibration mean so that the low-rank operator models only zero-mean variation, while $b_\ell$ restores the target mean at the output.
Finally, $U_\ell$ captures structured cross-feature interactions that the diagonal gate cannot represent, in a low-rank form, motivated by intrinsic dimensionality results for Transformer adaptation~\citep{aghajanyan-etal-2021-intrinsic, hu2022lora}; the overall linear form reflects the empirical near-linearity of per-layer transitions in Transformer decoders~\citep{razzhigaev-etal-2024-transformer}. 
Because $\widetilde{F}_\ell$ matches the input/output signature of $F_\ell$, the surrounding Transformer block is left unchanged~\citep{replaceme}. 
The two submodule types differ only in (i) the supervision target $y_i = F_\ell(x_i)$ and (ii) whether the rank-$r$ input basis $V_\ell$ is layer-specific or shared across layers.

\noindent\textbf{Closed-form rank-$r$ fit.}
The surrogate in Eq.~\eqref{eq:generic_surrogate} admits a closed-form fitting procedure on calibration statistics, with no gradient-based optimization. Setting $\mu_\ell$ to the empirical input mean and $b_\ell$ to the empirical target mean decouples the affine and low-rank components
(Frisch--Waugh--Lovell theorem): with $(\mu_\ell, b_\ell)$ fixed, the residual problem in $(g_\ell, U_\ell)$ is expressed entirely on centered quantities, and the two terms in Eq.~\eqref{eq:generic_surrogate} no longer share their constant component. The element-wise gate is recovered per feature as the ridge-regularized univariate slope
\begin{equation}
g_\ell^{(j)} \;=\;
\frac{\widetilde{\sigma}_{xy}^{(j)}}
     {\widetilde{\sigma}_{xx}^{(j)} + \lambda},
\label{eq:gate}
\end{equation}
where $\widetilde{\sigma}_{xx}^{(j)}$ and $\widetilde{\sigma}_{xy}^{(j)}$ are the centered input variance and input-target covariance on coordinate $j$, respectively, and $\lambda > 0$ is a small ridge term. Conditioned on $(g_\ell, b_\ell, \mu_\ell)$, the low-rank operator $U_\ell = V_\ell^\top W_\ell$ is fit in two steps. 
The input basis $V_\ell$ is set to the top-$r$ eigenvectors of the centered input covariance $\widetilde{\Sigma}_{xx}^{(\ell)} = \widetilde{X}_\ell^\top \widetilde{X}_\ell$, which selects the $r$-dimensional subspace of inputs along which the calibration activations carry most of their variance; this leverages the intrinsic low-dimensionality of Transformer activations~\citep{aghajanyan-etal-2021-intrinsic, hu2022lora}.
Conditioned on $V_\ell$, the output factor $W_\ell$ is then recovered by ridge-regularized least squares of the affine-corrected target onto the projection of the centered input on $V_\ell$. The full fit thus reduces, per submodule, to one accumulation of input covariance and cross-covariance statistics, one symmetric eigen-decomposition of size $d \times d$, and one rank-$r$ ridge regression.

\noindent\textbf{Attention replacement (independent bases).}
For a selected Attention submodule, the target $y_i$ is the dense Attention output after the architecture-specific output projection.
We instantiate Eq.~\eqref{eq:generic_surrogate} with a per-layer factorization $U_\ell^{\mathrm{attn}} = V_\ell^\top W_\ell$, so each removed Attention submodule owns its full parameter set $\theta_\ell^{\mathrm{attn}}=\{g_\ell,b_\ell,\mu_\ell,V_\ell,W_\ell\}$.
This reflects the fact that Attention contributions across depths perform heterogeneous roles~\citep{he2024matterstransformersattentionneeded,zhong-etal-2025-blockpruner}.

\noindent\textbf{FeedForward replacement (shared input basis).}
For a selected FFN submodule, the target $y_i$ is the dense residual output after projection back to the hidden size. 
A na\"ive instantiation of Eq.~\eqref{eq:generic_surrogate} would assign an independent basis $V_\ell$ to each removed submodule.
Instead, \methodname ties the input basis across $\ell \in \mathcal{S}_{\mathrm{fn}}$, setting $U_\ell^{\mathrm{fn}} = V^\top W_\ell$ with a single $V \in \mathbb{R}^{r \times d}$ shared across selected layers; only $\theta_\ell^{\mathrm{fn}} = \{g_\ell, b_\ell, \mu_\ell, W_\ell\}$ remains layer-specific. 
This realizes a cross-layer parameter-sharing structure~\citep{mikaelyan2025deltallmcompressllmslowrank}, i.e., a common $r$-dimensional input subspace across the removed submodules, with per-layer output coefficients $W_\ell$ specializing the map. 
We estimate the shared-basis surrogate from joint calibration statistics:
\begin{equation}
\min_{V,\,\{\theta_\ell\}_{\ell \in \mathcal{S}_{\mathrm{fn}}}} \;
\sum_{\ell \in \mathcal{S}_{\mathrm{fn}}}
\sum_{i=1}^{N}
\bigl\| \widetilde{F}^{\mathrm{fn}}_\ell\bigl(x_i^{(\ell)}\bigr)
- y_i^{(\ell)} \bigr\|_2^2 .
\label{eq:fn_shared_fit}
\end{equation}

In practice, $V$ is set to the top-$r$ eigenvectors of the sum of centered input covariances across all selected FFN layers; 
each $W_\ell$ is then recovered by an independent ridge least-squares. The shared basis is thus estimated once from joint statistics, while output coefficients remain layer-specific.

A second source of parameter savings comes from the surrogate operating directly on the residual stream of dimension $d$, bypassing the larger inner dimension $d_{ff} \gg d$ of the FFN block. 
Since a SwiGLU \cite{shazeer2020glu} FFN comprises three $d \times d_{ff}$ projections (gate, up, down) per selected layer, the bypass stores $W_\ell \in \mathbb{R}^{r \times d}$ together with the three $d$-dimensional vectors $g_\ell, b_\ell, \mu_\ell$, totaling $rd + 3d$ parameters, plus the shared $V \in \mathbb{R}^{r \times d}$ amortized across all $|\mathcal{S}_{\mathrm{fn}}|$ selected layers. Compared to the $3\,d\,d_{ff}$ parameters of the original block, the layer-specific parameter count is roughly $10\times$ smaller at $r=d$ for typical decoder LLMs with $d_{ff} \approx 3.5d$.

\noindent\textbf{Deployment form.}
After fitting, each selected submodule is replaced by its corresponding residual bypass, which approximates only the net residual contribution of the original submodule rather than its internal computation. The original model weights outside the selected submodules are left unchanged. For FFNs, the shared basis is stored once and reused across selected layers, while the layer-specific affine terms and output factors are stored per layer. This gives \methodname a compact deployed form relative to the removed submodules, while preserving a separate residual approximation for each removed component (see \Cref{tab:params-full}).

\section{Experiments}
\label{sec:experiments}
In this section, we describe the experimental setup and the numerical results of \methodname relative to the selected baselines. 
We organize our evaluation around three research questions: 
\textbf{(RQ1)}: How does submodule-level replacement, like \methodname, compare to layer-level replacement on perplexity and downstream accuracy? 
\textbf{(RQ2)}: How does \methodname scale with sparsity?
and \textbf{(RQ3)}: How does \methodname translate into inference-time efficiency?

\noindent\textbf{Experimental Setup.}
We outline the key design choices and evaluation protocol below; full details on models, sparsity levels, datasets, metrics, and calibration are provided in \Cref{app:setup}.
Note that GSM8K is reported separately in \Cref{sec:gsm8k}, due to its known sensitivity to prompting~\citep{sclar2024prompt} and instability under compression of generative reasoning~\citep{shrestha2026limits}. 

We consider four post-training, calibration-only block-replacement baselines that share the two design constraints relaxed by \methodname: \textbf{Streamline (FFN)} and \textbf{Streamline (Layer)}~\citep{streamline}, which replace a contiguous block of selected layers with a trained feed-forward network or a full Transformer layer, respectively; \textbf{ReplaceMe (LS)} and \textbf{ReplaceMe (Cosine)}~\citep{replaceme}, which fit a single linear transformation for the contiguous removed blocks, folded into existing weights at deployment. All methods use the same calibration data: SlimPajama~\citep{slimpajama} for base models and SlimOrca~\citep{orca} for instruction-tuned models, 8k samples at sequence length $1024$. For \methodname, we use rank $256$ for Attention bypasses and rank $4096$ for the shared FFN basis (see \Cref{sec:ablations}). Calibration sample-size and sequence-length sensitivity are reported in \Cref{sec:calibration-sensitivity}.

\noindent\textbf{Motivation: granularity and contiguity.}
To empirically motivate the transition from contiguous full-layer replacement~\citep{streamline,replaceme} to non-contiguous submodule replacement. \Cref{tab:granularity-contiguity} compares three variants on Llama-3.2-3B at 25\% sparsity.  \emph{Single-contig.} replaces a contiguous layer span with a single surrogate; \emph{Submod.-contig.} fits separate Attention and FFN bypasses while restricting selection to a contiguous span; and \emph{Submod.-noncontig.} selects and replaces Attention and FFN submodules independently across depth, as our proposed method. Non-contiguous submodule replacement achieves the best C4, WikiText-2 (WT2), and geometric-mean perplexity, while remaining close to single contiguous replacement on Lambada. As this is a single-model diagnostic, we treat it as empirical motivation rather than evidence of universal superiority.

\begin{table}[!ht]
\label{tab:granularity-contiguity}
\centering
\setlength{\tabcolsep}{4.5pt}
\renewcommand{\arraystretch}{1.08}
\begin{adjustbox}{max width=\columnwidth}
\begin{tabular}{lcccc}
\toprule
Variant & Lambada $\downarrow$ & C4 $\downarrow$ & WT2 $\downarrow$ & Geom. $\downarrow$ \\
\midrule
Single-contig.
    & \textbf{37.02} & 25.74 & 18.65 & 26.10 \\
Submod.-contig.
    & 48.86 & 29.87 & 22.54 & 32.04 \\
\rowcolor{lightgreen}
Submod.-noncontig.
    & 37.60 & \textbf{22.58} & \textbf{16.69} & \textbf{24.20} \\
\bottomrule
\end{tabular}
\end{adjustbox}
\caption{Comparison of replacement granularity and contiguity on Llama-3.2-3B at 25\% sparsity.}
\end{table}

\begin{table}[!ht]
\label{tab:ppl-main}
\centering
\begin{adjustbox}{max width=\columnwidth}
\setlength{\tabcolsep}{4pt}
\begin{tabular}{l l c c c c}
\toprule
Sparsity & Method & Lambada $\downarrow$ & C4 $\downarrow$ & WT2 $\downarrow$ & PPL Deg. $\downarrow$ \\
\midrule

\multicolumn{6}{c}{\texttt{\textbf{Llama-3.2-3B}}} \\
\midrule
\rowcolor{lightgrey} Dense & & 20.15 & 11.07 & 7.82 & 1.00$\times$ \\
\midrule
 & Streamline (FFN) & 31.65 & 19.48 & 18.73 & 1.88$\times$ \\
 & Streamline (Layer) & 27.92 & 18.59 & 16.66 & 1.71$\times$ \\
12.5\% & ReplaceMe (LS) & 28.35 & 18.41 & 13.33 & 1.59$\times$ \\
 & ReplaceMe (Cosine) & 27.84 & 18.34 & 13.30 & 1.57$\times$ \\
\rowcolor{lightgreen} \cellcolor{white}  & \methodname & \textbf{27.25} & \textbf{15.96} & \textbf{10.81} & \textbf{1.39$\times$} \\
\cmidrule(lr){1-6}
 & Streamline (FFN) & 51.22 & 36.68 & 42.96 & 3.59$\times$ \\
 & Streamline (Layer) & 47.93 & 33.62 & 38.35 & 3.28$\times$ \\
25\% & ReplaceMe (LS) & 52.15 & 36.35 & 44.07 & 3.63$\times$ \\
 & ReplaceMe (Cosine) & 50.22 & 31.34 & 23.47 & 2.77$\times$ \\
\rowcolor{lightgreen} \cellcolor{white}  & \methodname & \textbf{39.17} & \textbf{23.42} & \textbf{17.86} & \textbf{2.11$\times$} \\
\midrule
\multicolumn{6}{c}{\texttt{\textbf{Qwen3-4B}}} \\
\midrule
\rowcolor{lightgrey} Dense & & 33.80 & 19.89 & 13.67 & 1.00$\times$ \\
\midrule
 & Streamline (FFN) & 46.78 & 28.83 & 25.28 & 1.55$\times$ \\
 & Streamline (Layer) & 43.86 & 27.16 & 22.92 & 1.48$\times$ \\
 & ReplaceMe (LS) & 48.74 & 30.86 & 24.55 & 1.59$\times$ \\
12.5\% & ReplaceMe (Cosine) & 45.40 & 28.56 & 20.70 & 1.43$\times$ \\
\rowcolor{lightgreen} \cellcolor{white}  & \methodname & \textbf{39.40} & \textbf{23.87} & \textbf{19.86} & \textbf{1.27$\times$} \\
\cmidrule(lr){1-6}
 & Streamline (FFN) & 448.64 & 329.13 & 1451.81 & 28.57$\times$ \\
 & Streamline (Layer) & 314.53 & 282.55 & 1264.41 & 23.04$\times$ \\
25\% & ReplaceMe (LS) & 151.88 & 120.50 & 201.71 & 7.38$\times$ \\
 & ReplaceMe (Cosine) & 132.37 & 93.01 & 115.85 & 5.37$\times$ \\
\rowcolor{lightgreen} \cellcolor{white}  & \methodname & \textbf{69.26} & \textbf{43.52} & \textbf{50.08} & \textbf{2.54$\times$} \\
\midrule
\multicolumn{6}{c}{\texttt{\textbf{Llama-3.1-8B}}} \\
\midrule
\rowcolor{lightgrey} Dense & & 17.78 & 9.36 & 6.24 & 1.00$\times$ \\
\midrule
 & Streamline (FFN) & 22.10 & 14.07 & 10.00 & 1.44$\times$ \\
 & Streamline (Layer) & \textbf{21.65} & 13.75 & 9.76 & 1.41$\times$ \\
12.5\% & ReplaceMe (LS) & 22.35 & 14.19 & 9.52 & 1.43$\times$ \\
 & ReplaceMe (Cosine) & 22.14 & 14.16 & 9.46 & 1.42$\times$ \\
\rowcolor{lightgreen} \cellcolor{white}  & \methodname & 24.24 & \textbf{13.57} & \textbf{8.63} & \textbf{1.40$\times$} \\
\cmidrule(lr){1-6}
 & Streamline (FFN) & 35.90 & 25.60 & 20.84 & 2.64$\times$ \\
 & Streamline (Layer) & 33.88 & 24.84 & 19.30 & 2.50$\times$ \\
25\% & ReplaceMe (LS) & 43.28 & 29.84 & 32.25 & 3.42$\times$ \\
 & ReplaceMe (Cosine) & 54.63 & 34.60 & 31.29 & 3.85$\times$ \\
\rowcolor{lightgreen} \cellcolor{white}  & \methodname & \textbf{33.09} & \textbf{19.30} & \textbf{14.23} & \textbf{2.06$\times$} \\
\midrule
\multicolumn{6}{c}{\texttt{\textbf{Qwen3-8B}}} \\
\midrule
\rowcolor{lightgrey} Dense & & 25.70 & 15.25 & 9.72 & 1.00$\times$ \\
\midrule
 & Streamline (FFN) & 37.52 & 23.84 & 20.65 & 1.69$\times$ \\
 & Streamline (Layer) & 36.30 & 22.78 & \textbf{15.88} & 1.51$\times$ \\
12.5\% & ReplaceMe (LS) & 38.69 & 25.96 & 24.99 & 1.87$\times$ \\
 & ReplaceMe (Cosine) & 37.05 & 24.83 & 21.37 & 1.73$\times$ \\
\rowcolor{lightgreen} \cellcolor{white}  & \methodname & \textbf{34.32} & \textbf{20.13} & 17.03 & \textbf{1.46$\times$} \\
\cmidrule(lr){1-6}
 & Streamline (FFN) & 356.48 & 348.07 & 1332.92 & 35.14$\times$ \\
 & Streamline (Layer) & 229.51 & 204.43 & 855.48 & 21.92$\times$ \\
25\% & ReplaceMe (LS) & 112.81 & 89.45 & 177.44 & 7.77$\times$ \\
 & ReplaceMe (Cosine) & 102.44 & 77.96 & 150.20 & 6.80$\times$ \\
\rowcolor{lightgreen} \cellcolor{white}  & \methodname & \textbf{63.22} & \textbf{40.06} & \textbf{48.52} & \textbf{3.18$\times$} \\
\midrule
\multicolumn{6}{c}{\texttt{\textbf{DeepSeek-7B}}} \\
\midrule
\rowcolor{lightgrey} Dense & & 14.36 & 9.75 & 6.85 & 1.00$\times$ \\
\midrule
 & Streamline (FFN) & 20.37 & 15.16 & 12.23 & 1.58$\times$ \\
 & Streamline (Layer) & 19.33 & 14.47 & 10.77 & 1.46$\times$ \\
12.5\% & ReplaceMe (LS) & 18.75 & 14.26 & 11.09 & 1.46$\times$ \\
 & ReplaceMe (Cosine) & \textbf{18.21} & \textbf{14.12} & 10.65 & \textbf{1.42$\times$} \\
\rowcolor{lightgreen} \cellcolor{white}  & \methodname & 23.17 & 14.47 & \textbf{9.95} & 1.52$\times$ \\
\cmidrule(lr){1-6}
 & Streamline (FFN) & 45.88 & 38.77 & 39.20 & 4.17$\times$ \\
 & Streamline (Layer) & 35.37 & 29.38 & 26.06 & 3.05$\times$ \\
25\% & ReplaceMe (LS) & 34.35 & 29.64 & 28.42 & 3.11$\times$ \\
 & ReplaceMe (Cosine) & 41.79 & 35.28 & 40.95 & 3.98$\times$ \\
\rowcolor{lightgreen} \cellcolor{white}  & \methodname & \textbf{31.35} & \textbf{21.44} & \textbf{18.94} & \textbf{2.37$\times$} \\
\bottomrule
\end{tabular}
\end{adjustbox}
\caption{Perplexity comparison at representative sparsity levels. Lower PPL is better; PPL Deg. is the geometric mean of sparse/dense PPL ratios.}
\end{table}

\begin{table*}[t]
\centering
\label{tab:instruct-s025}
\begin{adjustbox}{max width=\textwidth}
\begin{tabular}{l c c c c c c c c c c c c c | c c}
\toprule
\multicolumn{2}{c}{} & \multicolumn{5}{c}{Commonsense} & \multicolumn{2}{c}{QA} & \multicolumn{1}{c}{Knowledge} & \multicolumn{2}{c}{Reasoning} & \multicolumn{1}{c}{Math} & \multicolumn{1}{c}{Science} & \multicolumn{2}{c}{Average} \\
\cmidrule(lr){3-7} \cmidrule(lr){8-9} \cmidrule(lr){10-10} \cmidrule(lr){11-12} \cmidrule(lr){13-13} \cmidrule(lr){14-14} \cmidrule(lr){15-16}
Model & Sparsity & HS & PIQA & Wino & OBQA & SIQA & BoolQ & TQA & MMLU & ARC-C & ARC-E & MathQA & SciQ & Avg & Acc. Ret. \\
\midrule

\rowcolor{lightgrey}\texttt{\textbf{Llama-3.2-3B-Instruct}} & Dense & 64.53 & 75.63 & 67.88 & 42.00 & 46.11 & 75.63 & 33.54 & 61.84 & 45.31 & 68.10 & 26.03 & 87.90 & 57.87 & 100.00\% \\
Streamline (FFN) &  & 50.25 & 64.58 & 67.09 & 30.80 & 41.30 & 76.42 & 28.89 & 34.45 & 33.62 & 47.39 & 21.78 & 68.30 & 47.07 & 81.33\% \\
Streamline (Layer) &  & 50.93 & 65.40 & 66.38 & 30.00 & 41.56 & 79.24 & 29.62 & 55.05 & 34.90 & 48.23 & 21.57 & 71.40 & 49.52 & 85.57\% \\
ReplaceMe (LS) & 25\% & 48.84 & 63.71 & 65.51 & 32.60 & 40.48 & 79.20 & 27.66 & 61.31 & 30.97 & 49.16 & \textbf{22.48} & 75.00 & 49.74 & 85.95\% \\
ReplaceMe (Cosine) &  & 50.45 & \textbf{65.51} & \textbf{67.25} & 32.20 & 40.79 & \textbf{79.30} & 30.35 & \textbf{61.35} & 31.91 & 47.60 & 22.28 & 75.80 & 50.40 & 87.08\% \\
\rowcolor{lightgreen}\methodname &  & \textbf{51.42} & \textbf{65.51} & 65.35 & \textbf{34.20} & \textbf{41.97} & 77.13 & \textbf{31.33} & 60.69 & \textbf{34.98} & \textbf{51.60} & 22.18 & \textbf{80.00} & \textbf{51.36} & \textbf{88.75\%} \\

\midrule

\rowcolor{lightgrey}\texttt{\textbf{Qwen3-4B-Instruct}} & Dense & 43.42 & 69.70 & 56.75 & 40.80 & 46.32 & 84.83 & 39.17 & 72.18 & 43.17 & 55.43 & 33.30 & 67.50 & 54.38 & 100.00\% \\
Streamline (FFN) &  & 22.93 & 49.95 & 42.23 & 33.80 & 34.54 & 62.26 & 24.60 & \textbf{71.74} & 26.28 & 27.61 & 25.26 & 22.80 & 37.00 & 68.04\% \\
Streamline (Layer) &  & 22.63 & 54.95 & \textbf{57.62} & 31.20 & 39.00 & 62.17 & 32.19 & 59.23 & \textbf{31.57} & 40.28 & 25.49 & 37.00 & 41.11 & 75.60\% \\
ReplaceMe (LS) & 25\% & \textbf{37.05} & \textbf{66.43} & 52.88 & \textbf{37.20} & 38.59 & 50.61 & \textbf{34.88} & 31.05 & 31.40 & \textbf{41.71} & \textbf{25.63} & 51.60 & 41.59 & 76.47\% \\
ReplaceMe (Cosine) &  & 36.72 & 66.32 & 54.30 & 35.20 & 38.69 & 52.11 & 33.66 & 31.56 & 29.78 & 40.36 & 24.25 & 49.80 & 41.06 & 75.51\% \\
\rowcolor{lightgreen}\methodname &  & 28.88 & 55.01 & 54.93 & 33.20 & \textbf{40.02} & \textbf{63.64} & 34.64 & 64.23 & 29.01 & 36.03 & 24.69 & \textbf{60.90} & \textbf{43.76} & \textbf{80.48\%} \\

\midrule

\rowcolor{lightgrey}\texttt{\textbf{Qwen2.5-7B-Instruct}} & Dense & 65.48 & 73.50 & 61.01 & 44.00 & 45.75 & 85.81 & 47.49 & 73.51 & 43.34 & 50.84 & 34.17 & 55.40 & 56.69 & 100.00\% \\
Streamline (FFN) &  & 51.87 & 71.93 & 52.88 & 39.00 & 40.84 & 60.31 & 28.76 & 27.67 & 33.96 & \textbf{44.78} & 27.71 & \textbf{56.30} & 44.67 & 78.79\% \\
Streamline (Layer) &  & 53.10 & \textbf{72.36} & 56.51 & 40.80 & 41.45 & 59.91 & 27.30 & 28.11 & 34.98 & 44.07 & 27.47 & 54.80 & 45.07 & 79.50\% \\
ReplaceMe (LS) & 25\% & 52.47 & 70.73 & 56.35 & 39.80 & 41.66 & 65.05 & 28.52 & 31.80 & \textbf{35.67} & 42.89 & \textbf{29.58} & 50.50 & 45.42 & 80.11\% \\
ReplaceMe (Cosine) &  & \textbf{53.17} & 71.49 & 56.51 & \textbf{41.40} & \textbf{42.48} & 65.20 & 29.25 & 32.86 & 34.56 & 42.34 & 28.44 & 51.60 & \textbf{45.77} & \textbf{80.74\%} \\
\rowcolor{lightgreen}\methodname &  & 48.19 & 69.26 & \textbf{57.85} & 40.20 & 41.91 & \textbf{68.10} & \textbf{33.66} & \textbf{35.59} & 34.13 & 41.79 & 27.84 & 49.40 & 45.66 & 80.54\% \\

\midrule

\rowcolor{lightgrey}\texttt{\textbf{DeepSeek-7B-chat}} & Dense & 70.56 & 77.64 & 74.90 & 44.80 & 50.46 & 83.98 & 37.21 & 50.90 & 41.55 & 62.63 & 27.54 & 78.40 & 58.38 & 100.00\% \\
Streamline (FFN) &  & 39.83 & 61.48 & 65.35 & 32.80 & 38.38 & 63.46 & \textbf{30.84} & 39.77 & \textbf{34.98} & 39.48 & 20.94 & 38.50 & 42.15 & 72.20\% \\
Streamline (Layer) &  & 50.02 & 66.32 & \textbf{70.95} & \textbf{35.60} & 44.37 & 64.01 & 27.54 & 48.68 & 31.14 & 44.07 & 21.88 & 58.50 & 46.92 & 80.37\% \\
ReplaceMe (LS) & 25\% & 47.15 & 63.77 & 70.88 & 35.20 & 43.50 & 75.84 & 29.25 & 48.94 & 32.51 & 42.72 & 22.81 & 48.20 & 46.73 & 80.05\% \\
ReplaceMe (Cosine) &  & \textbf{50.70} & \textbf{71.60} & 55.41 & 35.40 & 41.56 & 59.48 & 27.05 & 25.46 & 27.90 & \textbf{47.98} & 24.22 & \textbf{73.50} & 45.02 & 77.12\% \\
\rowcolor{lightgreen}\methodname &  & 48.64 & 65.02 & 70.72 & 34.00 & \textbf{46.37} & \textbf{83.09} & 29.74 & \textbf{49.96} & 33.11 & 46.13 & \textbf{24.62} & 58.40 & \textbf{49.15} & \textbf{84.19\%} \\

\midrule

\rowcolor{lightgrey}\texttt{\textbf{Llama-3.1-8B-Instruct}} & Dense & 72.52 & 79.49 & 77.82 & 49.00 & 50.20 & 83.88 & 40.39 & 68.79 & 53.58 & 75.76 & 26.80 & 91.70 & 64.16 & 100.00\% \\
Streamline (FFN) &  & 58.31 & 68.82 & 74.35 & 34.80 & 43.71 & \textbf{84.59} & 33.05 & 67.54 & 38.48 & 54.71 & 22.35 & 77.40 & 54.84 & 85.47\% \\
Streamline (Layer) &  & 59.90 & \textbf{70.46} & \textbf{75.22} & 34.00 & 44.01 & 84.56 & 33.90 & 67.77 & \textbf{40.19} & 56.82 & 22.18 & 80.20 & 55.77 & 86.92\% \\
ReplaceMe (LS) & 25\% & 57.27 & 67.90 & 72.61 & 35.60 & 44.01 & 84.56 & 34.39 & \textbf{68.00} & 38.74 & 54.67 & 21.24 & 75.50 & 54.54 & 85.01\% \\
ReplaceMe (Cosine) &  & \textbf{60.22} & 70.08 & 72.85 & 37.20 & 45.34 & 84.37 & 35.99 & 67.95 & 40.02 & 57.41 & 21.61 & 76.00 & 55.75 & 86.89\% \\
\rowcolor{lightgreen}\methodname &  & 59.33 & 69.48 & \textbf{75.22} & \textbf{39.20} & \textbf{45.96} & 84.43 & \textbf{36.23} & 67.27 & 39.93 & \textbf{59.85} & \textbf{23.22} & \textbf{84.10} & \textbf{57.02} & \textbf{88.87\%} \\

\bottomrule
\end{tabular}
\end{adjustbox}
\caption{Downstream performance (\%) at sparsity 25\%. Acc. Ret. is the ratio sparse/dense aggregate accuracy.}
\end{table*}

\noindent\textbf{RQ1: Perplexity and accuracy at the submodule-level.}
We analyze how \methodname is positioned on the perplexity–accuracy trade-off relative to the four replacement-based baselines.
Base-model perplexity (\Cref{tab:ppl-main}) and instruction-tuned downstream accuracy (\Cref{tab:instruct-s025}) report the two axes at representative sparsity levels; the remaining sparsities are deferred to \Cref{app:ppl} and \Cref{app:instruct}.

Considering perplexity, \methodname attains the lowest PPL degradation in 9 of 10 model–sparsity settings reported in \Cref{tab:ppl-main}, and is consistently strongest at $25\%$ sparsity. 
The separation is most pronounced on the Qwen family, where contiguous replacement baselines degrade sharply: at $25\%$ sparsity, the strongest ReplaceMe baseline reaches $5.37\times$ on Qwen3-4B and $6.80\times$ on Qwen3-8B, while \methodname reduces these to $2.54\times$ and $3.18\times$ respectively. 
On the Llama and DeepSeek families, the ordering is the same, with \methodname consistently best at $25\%$ sparsity.

On the accuracy side, \methodname obtains the best aggregate downstream accuracy on 4 of 5 models at $25\%$ sparsity (\Cref{tab:instruct-s025}), retaining $84.6\%$ of dense performance on average. 
The single exception is Qwen2.5-7B-Instruct, where ReplaceMe Cosine is marginally higher in aggregate ($+0.11\%$); even there, \methodname improves several individual tasks, including Winogrande~\cite{sakaguchi2021winogrande}, BoolQ~\cite{clark2019boolq}, TruthfulQA~\cite{lin2022truthfulqa}, and MMLU~\cite{hendrycks2021mmlu}.
At $25\%$ sparsity, it provides the strongest aggregate trade-off between base-model PPL degradation and instruction-tuned accuracy retention (\Cref{fig:teaser}).

\noindent\textbf{RQ2: Scaling with compression.}
\Cref{tab:summary-retention-ppl} aggregates accuracy retention over instruction-tuned models and PPL degradation over base models across all five sparsity levels. At the aggregate level, \methodname attains the strongest score on both axes at every evaluated sparsity level, and the margin over the strongest competing baseline widens monotonically with compression: the PPL gap grows from $0.11\times$ at $12.5\%$ to $1.92\times$ at $25\%$ and $5.69\times$ at $37.5\%$, while accuracy retention separates by $+0.2$, $+3.0$, and $+3.0$ points at the same sparsities. 
\methodname is also the only method to retain above $80\%$ accuracy at $25\%$ and above $73\%$ at $37.5\%$. Stability follows the same pattern: PPL-factor standard deviation stays bounded for \methodname ($\pm 0.08 \to \pm 1.37$), whereas baselines grow by one to two orders of magnitude (e.g.\ ReplaceMe Cosine reaches $\pm 130.68$ at $37.5\%$). 
These results suggest that non-contiguous submodule selection can extend the viable compression range, rather than merely improve a fixed operating point.

\begin{table}[!ht]
\label{tab:summary-retention-ppl}
\centering
\setlength{\tabcolsep}{4.5pt}
\renewcommand{\arraystretch}{1.08}
\begin{adjustbox}{max width=\columnwidth}
\begin{tabular}{lccccc}
\toprule
\multicolumn{6}{c}{Accuracy retention (\%) $\uparrow$} \\
\midrule
Method & 12.5\% & 20\% & 25\% & 30\% & 37.5\% \\
\midrule
Streamline (FFN) & 91.7\std{3.1} & 86.2\std{4.3} & 77.2\std{6.3} & 72.3\std{7.9} & 64.8\std{5.8} \\
Streamline (Layer) & 93.2\std{2.3} & 85.4\std{3.5} & 81.6\std{4.1} & 74.9\std{7.0} & 68.7\std{6.4} \\
ReplaceMe (LS) & 92.7\std{1.8} & 85.8\std{4.2} & 81.5\std{3.5} & 77.7\std{4.7} & 70.8\std{5.6} \\
ReplaceMe (Cosine) & 93.2\std{2.4} & 86.8\std{4.0} & 81.5\std{4.8} & 79.1\std{4.0} & 70.9\std{5.6} \\
\rowcolor{lightgreen}\methodname & \textbf{93.4\std{2.7}} & \textbf{88.1\std{3.4}} & \textbf{84.6\std{3.7}} & \textbf{80.8\std{3.8}} & \textbf{73.9\std{2.9}} \\
\midrule
\multicolumn{6}{c}{PPL factor ($\times$ dense) $\downarrow$} \\
\midrule
Method & 12.5\% & 20\% & 25\% & 30\% & 37.5\% \\
\midrule
Streamline (FFN) & 1.62\std{.15} & 2.43\std{.70} & 8.31\std{14.07} & 7.06\std{3.09} & 16.01\std{14.17} \\
Streamline (Layer) & 1.52\std{.11} & 2.50\std{.61} & 6.61\std{9.58} & 6.61\std{4.96} & 12.44\std{13.21} \\
ReplaceMe (LS) & 1.58\std{.16} & 2.82\std{1.17} & 4.67\std{2.06} & 8.09\std{5.86} & 11.45\std{10.01} \\
ReplaceMe (Cosine) & 1.51\std{.12} & 2.46\std{.75} & 4.34\std{1.40} & 7.38\std{7.99} & 22.28\std{130.68} \\
\rowcolor{lightgreen}\methodname & \textbf{1.40\std{.08}} & \textbf{1.81\std{.22}} & \textbf{2.42\std{.40}} & \textbf{3.22\std{.51}} & \textbf{5.76\std{1.37}} \\
\bottomrule
\end{tabular}
\end{adjustbox}
\caption{Aggregate compression trade-off across sparsity levels (avg\std{std}). Accuracy retention is averaged over instruction-tuned models; PPL factor is averaged over base models, with each model-level factor computed from Lambada, C4, and WikiText-2.}
\end{table}

\noindent\textbf{RQ3: Inference efficiency.}
\Cref{tab:inference-speed-kv} reports inference diagnostics comparing dense and \methodname-compressed variants on a single NVIDIA A100 80GB GPU. At $25\%$ sparsity, time-to-first-token (TTFT) speedup over the dense model ranges from $1.18\times$ (Llama-3.2-3B) to $1.40\times$ (DeepSeek-7B), with decode speedup between $1.12\times$ and $1.17\times$. As expected, KV-cache usage scales proportionally with sparsity, due to the complete removal of Attention submodules. These gains are consistent with the reduction in active Attention and FFN submodules, without hardware-specific optimization. A matched comparison with the replacement-based baselines, including deployed overhead and realized generation speed, is provided in \Cref{tab:matched-baseline-efficiency}.

\begin{table}[!ht]
\label{tab:inference-speed-kv}
\centering
\setlength{\tabcolsep}{4pt}
\begin{adjustbox}{width=1\columnwidth}
\begin{tabular}{llrrr}
\toprule
Model & Sparsity & TTFT$\uparrow$ & Decode$\uparrow$ & KV-cache (MB, \% saved) \\
\midrule
\multirow{3}{*}{\texttt{\textbf{Llama-3.2-3B}}}
 & 12.5\% & 1.13$\times$ & 1.06$\times$ & 56$\to$48 (14.3\%) \\
 & 25.0\% & 1.18$\times$ & 1.12$\times$ & 56$\to$42 (25.0\%) \\
 & 37.5\% & 1.22$\times$ & 1.18$\times$ & 56$\to$36 (35.7\%) \\
\midrule
\multirow{3}{*}{\texttt{\textbf{{Qwen3-4B}}}} 
 & 12.5\% & 1.10$\times$ & 1.06$\times$ & 72$\to$64 (11.1\%) \\
 & 25.0\% & 1.18$\times$ & 1.17$\times$ & 72$\to$54 (25.0\%) \\
 & 37.5\% & 1.30$\times$ & 1.27$\times$ & 72$\to$44 (38.9\%) \\
\midrule
\multirow{3}{*}{\texttt{\textbf{Llama-3.1-8B}}}
 & 12.5\% & 1.17$\times$ & 1.06$\times$ & 64$\to$56 (12.5\%) \\
 & 25.0\% & 1.26$\times$ & 1.13$\times$ & 64$\to$48 (25.0\%) \\
 & 37.5\% & 1.33$\times$ & 1.20$\times$ & 64$\to$40 (37.5\%) \\
\midrule
\multirow{3}{*}{\texttt{\textbf{Qwen3-8B}}} 
 & 12.5\% & 1.15$\times$ & 1.07$\times$ & 72$\to$64 (11.1\%) \\
 & 25.0\% & 1.24$\times$ & 1.17$\times$ & 72$\to$54 (25.0\%) \\
 & 37.5\% & 1.37$\times$ & 1.29$\times$ & 72$\to$44 (38.9\%) \\
 \midrule
 \multirow{3}{*}{\texttt{\textbf{DeepSeek-7B}}} 
 & 12.5\% & 1.31$\times$ & 1.05$\times$ & 240$\to$208 (13.3\%) \\
 & 25.0\% & 1.40$\times$ & 1.12$\times$ & 240$\to$176 (26.7\%) \\
 & 37.5\% & 1.49$\times$ & 1.16$\times$ & 240$\to$152 (36.7\%) \\
\bottomrule
\end{tabular}
\end{adjustbox}
\caption{Inference speedup and KV-cache reduction. TTFT uses one generated token; decode uses 128 tokens; KV-cache is measured with a 512-token prompt.}
\end{table}

\noindent\textbf{Offline calibration cost.}
Since \methodname requires activation collection, submodule selection, and closed-form bypass fitting, we also report its calibration cost. \Cref{tab:calibration-cost} provides a stage-wise breakdown at $25\%$ sparsity on a single NVIDIA A100 80GB GPU. Calibration takes 25.1--32.1 minutes and requires 11.3--19.3 GB of peak GPU memory. FFN fitting accounts for most of the runtime due to the rank-$4096$ shared-basis regression. This offline cost does not affect inference latency.

\begin{table}[!ht]
\label{tab:calibration-cost}
\centering
\scriptsize
\setlength{\tabcolsep}{2.5pt}
\renewcommand{\arraystretch}{1.08}
\begin{adjustbox}{width=\columnwidth}
\begin{tabular}{l c c r c c}
\toprule
Model & Total & Selection & Attn. fit & FFN fit & Peak mem. \\
\midrule
\texttt{\textbf{Llama-3.2-3B}}
    & 25.2 min & 0.7 min & 6.5 min & 17.7 min & 11.3 GB \\
\texttt{\textbf{Qwen3-4B}}
    & 31.3 min & 0.7 min & 9.2 min & 21.1 min & 18.9 GB \\
\texttt{\textbf{Llama-3.2-3B-Inst.}}
    & 25.1 min & 0.3 min & 7.1 min & 17.4 min & 11.6 GB \\
\texttt{\textbf{Qwen3-4B-Inst.}}
    & 32.1 min & 0.5 min & 10.0 min & 21.3 min & 19.3 GB \\
\bottomrule
\end{tabular}
\end{adjustbox}
\caption{Offline calibration cost of \methodname at $25\%$ sparsity on an A100 80GB GPU in BF16, using 8k calibration sequences of length 1024. Total time includes activation collection, scoring, and bypass fitting, but excludes model loading and downstream evaluation.}
\end{table}

\section{Ablation Studies}
\label{sec:ablations}

To analyze the contribution of each design choice, we conduct five ablation studies at 25\% sparsity on Llama-3B and Qwen3-4B, reporting perplexity on Lambada, C4, and WikiText-2 for base models and a reduced five-task subset for instruction-tuned models (with \emph{Inst.\ Avg.} interpreted as a relative indicator within this section).

\noindent\textbf{Residual compensation.}
\Cref{tab:ablation-compensation-tasks} isolates the role of residual compensation.
\emph{Pruning only} removes the selected submodules without any replacement, leading to severe degradation, which confirms that na\"ive deletion is not viable at this sparsity.
Compensating Attention alone (\emph{w/ Attn.}) improves over pruning-only but still suffers high degradation, whereas compensating FFNs alone (\emph{w/ FeedForward}) recovers most of the degradation. This suggests that approximating removed FFN residuals is the dominant factor for stability in this setting.
The full \methodname configuration (\emph{w/ Attn.+FeedForward}) yields the best downstream accuracy on both models, with Qwen3-4B base PPL slightly favoring FeedForward-only compensation. This asymmetry motivates \methodname's design: per-layer parameters for Attention, and a shared input basis across FFNs. 

\begin{table}[!ht]
\label{tab:ablation-compensation-tasks}
\centering
\setlength{\tabcolsep}{4.0pt}
\renewcommand{\arraystretch}{1.1}
\begin{adjustbox}{width=\columnwidth}
\begin{tabular}{llcccc}
\toprule
 & & \multicolumn{3}{c}{Base PPL $\downarrow$} & Inst. Avg. $\uparrow$ \\
\cmidrule(lr){3-5}
Model & Setup & LAMB. & C4 & WT2 & \\
\midrule
\multirow{4}{*}{\texttt{\textbf{Llama-3B}}} & Pruning only & $1.30\times 10^{4}$ & $1.01\times 10^{4}$ & $1.12\times 10^{4}$ & 60.8 \\
 & w/ Attn. comp. & $5.32\times 10^{3}$ & $9.52\times 10^{3}$ & $1.11\times 10^{4}$ & 62.2 \\
 & w/ FeedForward comp. & \textbf{37.42} & 22.94 & 16.98 & 67.7 \\
\rowcolor{lightgreen} \cellcolor{white}  & w/ Attn.+FeedForward \textbf{(Ours)}  & 37.82 & \textbf{22.26} & \textbf{16.57} & \textbf{69.5} \\
\midrule
\multirow{4}{*}{\texttt{\textbf{Qwen3-4B}}} & Pruning only & 212 & 92.33 & 79.27 & 50.9 \\
 & w/ Attn. comp. & 82.92 & 53.58 & 67.73 & 48.0 \\
 & w/ FeedForward comp. & 74.60 & 48.13 & \textbf{39.71} & 55.6 \\
\rowcolor{lightgreen} \cellcolor{white}  & w/ Attn.+FeedForward \textbf{(Ours)}  & \textbf{69.36} & \textbf{43.49} & 50.10 & \textbf{56.9} \\
\bottomrule
\end{tabular}
\end{adjustbox}
\caption{Residual compensation component ablation at 25\% sparsity without cross-task aggregation. Base rows report raw task-wise PPL; instruction-tuned rows report raw downstream average accuracy. Large PPL values are shown in scientific notation.}
\end{table}

\noindent\textbf{Shared versus per-layer FeedForward basis.}
\Cref{tab:ablation-FeedForward-basis} compares the shared FFN basis used by \methodname against a per-layer basis variant, applied to the same selected FFN layers. 
Across all base-model PPL metrics, differences are at most $0.01$ in raw perplexity, and instruction-tuned averages differ by at most $0.2$ accuracy points.
At the same time, the per-layer variant stores an independent $V_\ell \in \mathbb{R}^{r \times d}$ for each removed layer, while the shared variant stores a single $V$ reused across all selected FFNs, reducing the basis cost by a factor of $|\mathcal{S}_{\text{fn}}|$. 
The shared design, therefore, matches the per-layer variant while using much fewer deployed parameters.

\begin{table}[!ht]
\centering
\label{tab:ablation-FeedForward-basis}
\setlength{\tabcolsep}{4.0pt}
\renewcommand{\arraystretch}{1.1}
\begin{adjustbox}{width=1\columnwidth}
\begin{tabular}{llcccc}
\toprule
 & & \multicolumn{3}{c}{Base PPL $\downarrow$} & Inst. Avg. $\uparrow$ \\
\cmidrule(lr){3-5}
Model & Basis & LAMB. & C4 & WT2 & \\
\midrule
 & Per-layer & \textbf{33.55} & \textbf{19.95} & 14.01 & \textbf{72.7} \\
\rowcolor{lightgreen} 
\cellcolor{white} \multirow{-2}{*}{\texttt{\textbf{Llama-3B}}} & Shared \textbf{(Ours)} & \textbf{33.55} & 19.96 & \textbf{14.00} & 72.5 \\
\midrule
 & Per-layer & 50.01 & \textbf{28.95} & \textbf{19.11} & \textbf{66.7} \\
\rowcolor{lightgreen} 
\cellcolor{white} \multirow{-2}{*}{\texttt{\textbf{Qwen3-4B}}} & Shared \textbf{(Ours)} & \textbf{50.00} & 28.97 & 19.12 & 66.6 \\
\bottomrule
\end{tabular}
\end{adjustbox}
\caption{Shared versus per-layer FeedForward basis at $25\%$ sparsity under fixed selected layers.}
\end{table}

\begin{figure}[!ht]
    \centering
    \includegraphics[width=1\columnwidth]{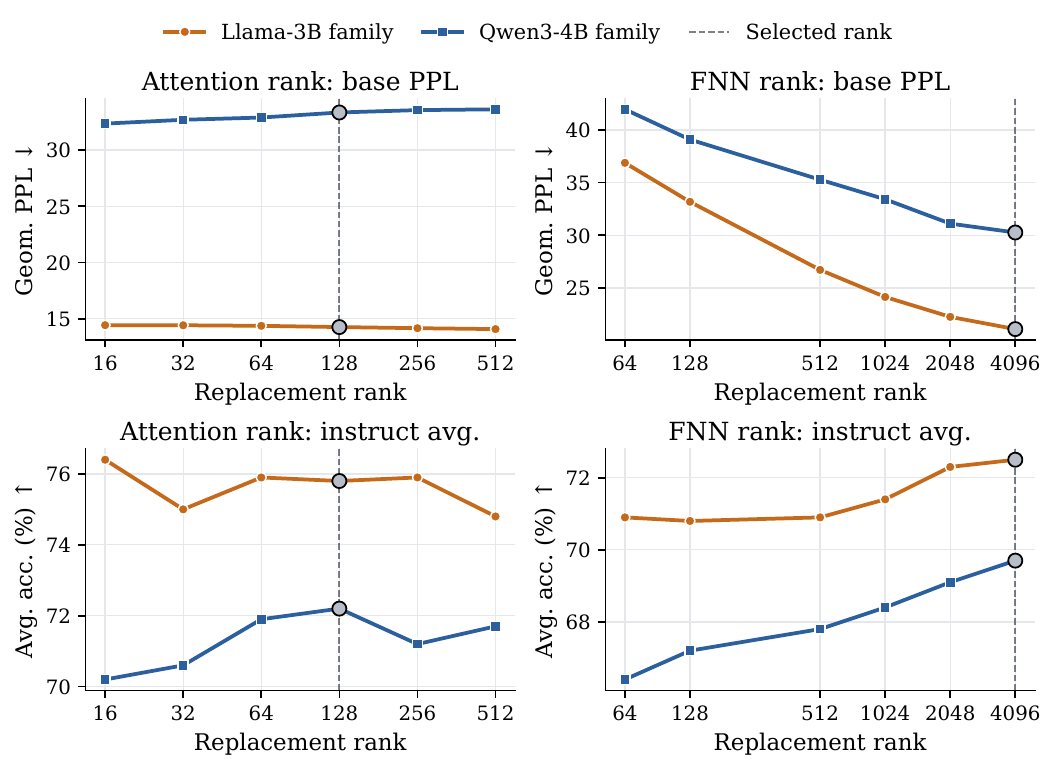}
    \caption{Fixed-selection rank sensitivity at $25\%$ sparsity. FFN quality improves steadily up to rank $4096$, while Attention saturates already at rank $256$, confirming that the two submodule types require different ranks.}
    \label{fig:rank-sensitivity}
    \vspace{-0.2cm}
\end{figure}

\noindent\textbf{Rank sensitivity.}
We sweep Attention and FFN bypass ranks independently, holding the selected layers fixed to isolate rank effects from layer-selection effects. As shown in \Cref{fig:rank-sensitivity}, the two submodule types exhibit qualitatively different behavior: FFN quality improves steadily and plateaus around rank $4096$, suggesting that a higher-dimensional shared input subspace is beneficial to capture the residual contribution of removed FFNs, whereas Attention saturates already at rank $256$, suggesting a low-rank operator suffices. \methodname therefore uses rank $256$ for Attention bypasses and rank $4096$ for the shared FFN basis, matching each submodule to the smallest rank past which gains plateau.

\noindent\textbf{Masking as a proxy for replacement quality.}
We compare deletion-only scoring with post-replacement scoring to test whether masked degradation is a reliable proxy for replacement quality. \Cref{tab:masked-replaced-proxy-direction} shows that the two criteria often select different scores: the criterion that minimizes masked degradation is not always the one that yields the best post-replacement outcome. This mismatch appears across both tested model families and submodule types, suggesting that replacement quality depends not only on the local effect of masking a submodule, but also on how well its residual contribution can be approximated by the surrogate.

\begin{table}[!ht]
\label{tab:masked-replaced-proxy-direction}
\centering
\setlength{\tabcolsep}{3.0pt}
\renewcommand{\arraystretch}{1.08}
\begin{adjustbox}{width=\columnwidth}
\begin{tabular}{llccc}
\toprule
Model & Submodule & Masking pref. & Replacement pref. & Match\\
\midrule
\multicolumn{5}{c}{\textit{Base models: WikiText-2 PPL $\downarrow$}} \\
\midrule
\texttt{\textbf{Llama-3B}}  & Attn. & Impact {\scriptsize (10.07)} & Impact {\scriptsize (9.50)}   & \checkmark \\
\texttt{\textbf{Llama-3B}}  & FeedForward   & Cosine {\scriptsize (45.28)} & Impact {\scriptsize (14.01)}  & $\times$ \\
\texttt{\textbf{Qwen3-4B}}  & Attn. & Impact {\scriptsize (36.66)} & Cosine {\scriptsize (24.11)} & $\times$ \\
\texttt{\textbf{Qwen3-4B}}  & FeedForward   & Impact {\scriptsize (23.56)} & Impact {\scriptsize (19.12)}  & \checkmark \\
\midrule
\multicolumn{5}{c}{\textit{Instruction-tuned: Avg. Acc. $\uparrow$}} \\
\midrule
\texttt{\textbf{Llama-3B-Inst.}} & Attn. & Cosine {\scriptsize (74.8)} & Cosine {\scriptsize (76.2)} & \checkmark \\
\texttt{\textbf{Llama-3B-Inst.}} & FeedForward   & Tie {\scriptsize (63.7)}     & Cosine {\scriptsize (73.0)}  & $\times$ \\
\texttt{\textbf{Qwen3-4B-Inst.}} & Attn. & Cosine {\scriptsize (69.4)} & Impact {\scriptsize (71.1)}  & $\times$ \\
\texttt{\textbf{Qwen3-4B-Inst.}} & FeedForward   & Tie {\scriptsize (58.4)}     & Cosine {\scriptsize (68.6)}  & $\times$ \\
\bottomrule
\end{tabular}
\end{adjustbox}
\caption{Masked degradation and post-replacement quality disagree in 5 out of 8 cases (marked ``$\times$''), indicating that masked degradation is not a reliable proxy for replacement quality.}
\end{table}

\noindent\textbf{Replacement-aware versus cosine FFN scoring.}
We compare replacement-aware and cosine FFN scoring to analyze how the selection criterion affects post-replacement quality, while keeping Attention selection and all replacement hyperparameters fixed. \Cref{tab:ffn-scoring-comparison} reports base-model perplexity and instruction-tuned downstream accuracy at 25\% sparsity. For base models, replacement-aware scoring achieves the lowest geometric-mean PPL on both model families, with a larger improvement on Qwen3-4B. On the other hand, cosine scoring obtains the highest instruction-tuned average on both models. These results are consistent with the final design of \methodname, which uses replacement-aware scoring for base-model FFNs and cosine scoring for instruction-tuned FFNs, while using replacement-aware Attention selection in both settings.

\begin{table}[!ht]
\label{tab:ffn-scoring-comparison}
\centering
\scriptsize
\setlength{\tabcolsep}{1.8pt}
\renewcommand{\arraystretch}{1.08}

\begin{tabularx}{\columnwidth}{@{}l c Y Y Y|Y@{}}
\toprule
\multicolumn{6}{c}{(a) Base-model PPL $\downarrow$} \\
\midrule
Model & Score & LAMB. & C4 & WT2 & Geom. \\
\midrule
\rowcolor{lightgreen}
\texttt{\textbf{Llama-3B}} & RA
    & 37.61 & \textbf{22.59} & \textbf{16.70}
    & \textbf{24.21} \\
\texttt{\textbf{Llama-3B}} & Cos.
    & \textbf{37.44} & 27.47 & 22.61
    & 28.54 \\
\rowcolor{lightgreen}
\texttt{\textbf{Qwen3-4B}} & RA
    & \textbf{69.10} & \textbf{44.53} & \textbf{50.87}
    & \textbf{53.89} \\
\texttt{\textbf{Qwen3-4B}} & Cos.
    & 93.42 & 67.45 & 100.35
    & 85.83 \\
\bottomrule
\end{tabularx}

\vspace{5pt}

\begin{tabularx}{\columnwidth}{@{}l c Y Y Y Y Y|Y@{}}
\toprule
\multicolumn{8}{c}{(b) Instruction-tuned accuracy (\%) $\uparrow$} \\
\midrule
Model & Score & PIQA & Wino & ARC-C & BoolQ & TQA & Avg. \\
\midrule
\texttt{\textbf{Llama-3B-Inst.}} & RA
    & \textbf{76.56} & 59.38 & \textbf{39.06}
    & 43.75 & 28.12 & 49.38 \\
\rowcolor{lightgreen}
\texttt{\textbf{Llama-3B-Inst.}} & Cos.
    & 62.50 & \textbf{73.44} & 31.25
    & \textbf{71.88} & \textbf{31.25} & \textbf{54.06} \\
\texttt{\textbf{Qwen3-4B-Inst.}} & RA
    & \textbf{56.25} & \textbf{57.81} & 29.69
    & 28.12 & 28.12 & 40.00 \\
\rowcolor{lightgreen}
\texttt{\textbf{Qwen3-4B-Inst.}} & Cos.
    & 45.31 & 53.12 & \textbf{31.25}
    & \textbf{73.44} & \textbf{32.81} & \textbf{47.19} \\
\bottomrule
\end{tabularx}
\caption{Replacement-aware (RA) versus cosine FFN scoring at 25\% sparsity. Panel (a) reports PPL; panel (b) reports instruction-tuned accuracy using 64 examples per task and is not directly comparable with the full downstream evaluation.}
\end{table}

\section{Conclusion}
\label{sec:conclusion}

This work asked whether the granularity at which replacement-based compression methods operate is itself a design choice that can be improved. Within the evaluated 3B--8B setting, our results suggest that this granularity can be improved: selecting Attention and FFN submodules independently across depths, and replacing each with a fitted residual bypass, recovers more of the dense model's behavior than contiguous full-layer replacement under the same parameter budget. 

The asymmetric nature of the recovery, where FFN compensation accounts for most of the recovered quality and Attention compensation refines the residual, also justifies the asymmetric design of \methodname: per-layer parameters for Attention, and a shared input basis for FFNs. The closed-form fit on calibration statistics makes the method cheap to deploy and easy to combine with existing inference stacks, with no sparse kernels or custom hardware required. The result that submodule-level recovery provides a stronger aggregate trade-off than layer-level recovery in our evaluated settings suggests that granularity is an underexplored axis in post-training compression, one that may generalize to other removal units such as attention heads or individual neurons.

\clearpage

\section*{Limitations}
\label{sec:limitations}

\methodname introduces explicit bypass parameters for selected Attention and FFN submodules, adding approximately $10\%$ of removed parameters and $15\%$ of removed MACs at $25\%$ sparsity (\Cref{tab:params-full} and \Cref{tab:macs-full} in the Appendix). 
This overhead is stable across sparsity levels. In contrast, ReplaceMe \cite{replaceme} variants add no deployed parameters as their transformations are folded into existing weights. In the matched benchmark reported in \Cref{tab:matched-baseline-efficiency}, \methodname provides smaller TTFT and decode speedups than the replacement-based baselines because its explicit residual bypasses introduce additional deployed computation. Its quality improvements therefore come with an explicit accuracy--efficiency trade-off.
The offline compression pipeline requires approximately 2{,}000--2{,}600 seconds on average, comparable to ReplaceMe Cosine but slower than Streamline \cite{streamline}, due to sequential selection and closed-form fitting. A stage-wise breakdown of calibration time and peak memory is reported in \Cref{tab:calibration-cost}, while the complete runtime comparison is given in \Cref{tab:runtime-summary}. This cost is incurred once during compression and does not affect deployed inference latency.

Our evaluation covers models between 3B and 8B parameters and therefore does not establish that the observed gains transfer to 32B- or 70B-scale models. A fair comparison at these scales would also require validated multi-GPU implementations of the replacement-based baselines, whose current implementations assume single-GPU execution. Extending \methodname and its baselines to larger models remains future work.

\section*{Ethical Considerations}
This work focuses on post-training compression for large language models. All models, datasets, and benchmarks used in our experiments are publicly available, and we credit the original authors throughout the manuscript. Our method does not introduce new training data or collect user data; it operates on pretrained models using calibration samples to fit lightweight replacement modules.
As with other model-compression techniques, \methodname can make LLM deployment cheaper and more accessible. This may broaden beneficial uses of language models, but it may also lower the cost of deploying models in harmful or unintended applications. Our work addresses the technical problem of deployable compression and does not target any specific downstream application.
Compression may also affect model behavior unevenly across tasks, domains, or demographic groups. To reduce the risk of hiding such effects, we report performance across multiple benchmarks, model families, and sparsity levels, and we separately discuss unstable settings such as GSM8K. However, our evaluation does not constitute a complete fairness or safety audit of compressed models. Further task-specific evaluation is needed before deploying compressed models in high-stakes settings.

\bibliography{custom}


\clearpage
\appendix
\startcontents[appendices]

\begin{center}
{\bfseries\large Appendix Table of Contents}
\end{center}

\begingroup
\setcounter{tocdepth}{2}
\printcontents[appendices]{}{1}{}
\endgroup

\setcounter{table}{0}
\renewcommand{\thetable}{A\arabic{table}}
\setcounter{figure}{0}
\renewcommand{\thefigure}{A\arabic{figure}}

\section{Evaluation Tasks}
\label{app:tasks}

\subsection{Perplexity datasets (base models).}
We evaluate perplexity on three standard benchmarks covering complementary text domains: Lambada (\citealp{paperno2016lambada}) for narrative text, C4 (\citealp{raffel2020t5}) for web text, and WikiText-2 (\citealp{merity2016pointer}) for encyclopedic text.

\subsection{Downstream tasks (instruction-tuned models).}
We evaluate on thirteen downstream tasks grouped by capability.
\emph{Commonsense reasoning:}
HellaSwag~\citep{zellers2019hellaswag},
PIQA~\citep{bisk2020piqa},
Winogrande~\citep{sakaguchi2021winogrande},
OpenBookQA~\citep{mihaylov2018openbookqa},
SocialIQA~\citep{sap2019socialiqa}.
\emph{Question answering:}
BoolQ~\citep{clark2019boolq},
TruthfulQA~\citep{lin2022truthfulqa}.
\emph{Knowledge:}
MMLU~\citep{hendrycks2021mmlu}.
\emph{Reasoning:}
ARC-Challenge and ARC-Easy~\citep{clark2018arc}.
\emph{Math:}
MathQA~\citep{amini2019mathqa},
GSM8K~\citep{cobbe2021gsm8k}.
\emph{Science:}
SciQ~\citep{welbl2017sciq}.
All tasks have been evaluated using a zero-shot setup apart from Winogrande, MMLU, and GSM8K, which have been evaluated in a five-shot scenario.

\subsection{Ablation subset.}
The ablation studies in \Cref{sec:ablations} use a reduced zero-shot subset of five tasks: PIQA, WinoGrande, ARC-Easy, BoolQ, and SciQ. All tasks use the LM Evaluation Harness~\citep{eval-harness}.

\section{Experimental Setup Details}
\label{app:setup}

\subsection{Models}
We evaluate \methodname on ten LLMs spanning three families and the 3B--8B scale. The five base models are Llama-3.2-3B \cite{grattafiori2024llama}, Qwen3-4B \cite{yang2025qwen3}, Llama-3.1-8B \cite{grattafiori2024llama}, Qwen3-8B \cite{yang2025qwen3}, and DeepSeek-7B \cite{bi2024deepseek}. The five instruction-tuned models are Llama-3.2-3B-Instruct \cite{grattafiori2024llama}, Qwen3-4B-Instruct \cite{yang2025qwen3}, Qwen2.5-7B-Instruct \cite{hui2024qwen2}, Llama-3.1-8B-Instruct \cite{grattafiori2024llama}, and DeepSeek-7B-chat \cite{bi2024deepseek}. All models are loaded from the HuggingFace Hub using their public checkpoints; we operate on the full-precision (bf16) weights and do not apply quantization at any stage of compression or evaluation.

\subsection{Sparsity levels}
We evaluate at five sparsity levels: $12.5\%$, $20\%$, $25\%$, $30\%$, and $37.5\%$. For a model with $L$ Transformer layers and target sparsity $s$, \methodname selects $\mathrm{round}(L \cdot s)$ Attention submodules and the same number of FFN submodules; the baselines remove $\mathrm{round}(L \cdot s)$ full Transformer layers, matching the overall parameter budget.

\subsection{Evaluation tasks}
For base models, we report perplexity on Lambada, C4, and WikiText-2, and summarize each model with the \emph{PPL factor}, defined as the geometric mean of sparse-to-dense PPL ratios across the three datasets, as done in~\cite{cunegatti2025zerothorder}. For instruction-tuned models, we report zero/few-shot accuracy on thirteen downstream tasks grouped into six categories: commonsense reasoning, question answering, knowledge, reasoning, math, and science. The complete task list with dataset references is in \Cref{app:tasks}. All downstream evaluations are run with the LM Evaluation Harness~\citep{eval-harness} under identical conditions across methods.

\subsection{Baselines}
We compare the proposed method against four block-replacement baselines that match \methodname's setting, i.e., post-training replacement of removed Transformer components using calibration data. Specifically, the four baselines are based on LLM-Streamline \cite{streamline} and ReplaceMe \cite{replaceme}. 

The former removes a contiguous sequence of layers and trains a lightweight replacement module to recover the lost computation. We include both (1) its FeedForward replacement variant (referred to as \textbf{Streamline (FFN)}), which selects a contiguous range of layers by cosine similarity between input and output residual states and replaces the block with a trained feed-forward network; and (2) its full layer replacement variant (referred to as \textbf{Streamline (Layer)}),  which uses the same selection criterion but substitutes the removed block with a full Transformer layer (self-attention plus feed-forward). 

ReplaceMe, on the other hand, removes a contiguous block of Transformer layers and fits a single linear map from calibration activations, which can be folded into adjacent weights at deployment. We include both (1) its least-squares variant (referred to as \textbf{ReplaceMe (LS)}); and (2) its cosine-based selection variant (referred to as \textbf{ReplaceMe (Cosine)}). 
The two variants differ in the criterion used to identify the contiguous range of Transformer layers: least-squares reconstruction error (LS), or cosine similarity (Cosine).

These four baselines therefore share the two design constraints relaxed by \methodname, i.e., full-layer granularity and contiguous selection, while remaining in the same post-training, calibration-based replacement setting.

Note that we do not include, among our baselines, LinearPatch~\citep{chen2026simple}, GRASP~\citep{liu2025grasp}, or ELM~\citep{jiang2026equilibrium} as main baselines because they target related but different recovery settings, even though we include them in the related work. Specifically, LinearPatch starts from an already-layer-pruned model and inserts a lightweight linear correction between the remaining blocks to reduce activation-magnitude mismatch; it does not jointly select and replace submodules from the dense model. GRASP is a layer-level hybrid pruning method: it identifies redundant layers and retains sensitivity-aware singular components of the original layer weights, rather than fitting a new residual map from calibration activations for each removed Attention or FeedForward submodule. Lastly, ELM replaces groups of Transformer layers with fixed-point modules selected by a learned policy and adapted with task-specific LoRA fine-tuning, whereas our setting is post-training and calibration-only. We therefore use Streamline and ReplaceMe as the main experimental comparisons, and discuss these other methods as related recovery approaches rather than direct baselines.

We also do not include any \emph{structured pruning} algorithm such as SliceGPT \cite{ashkboos2024slicegpt}, LLM-Pruner \cite{ma2023llm}, Laco \cite{yang2024laco}, UIDL \cite{gromov2025the} since the replacement-based compression literature has already shown to achieve better performance compared to standard structured pruning in \cite{streamline,replaceme}. This is also supported by our results in \Cref{tab:ablation-compensation-tasks}.

\subsection{Calibration Data}
All methods use the same calibration corpus and budget. For base models, we use SlimPajama~\citep{slimpajama}; for instruction-tuned models, we use SlimOrca~\citep{orca}. In both cases, we sample 8k sequences at maximum length $1024$ tokens, yielding approximately 8M calibration tokens per fit. No calibration data overlaps with any evaluation benchmark.

\subsection{\methodname hyperparameters}
For \methodname, we use attention bypass rank $r = 256$ and FFN shared-basis rank $r = 4096$; both ranks were selected based on the rank-sensitivity ablation (\Cref{sec:ablations}). The ridge regularization in the closed-form fit Eq.~\eqref{eq:gate} is set to $\lambda = 10^{-6}$; the small constant in the impact score in Eq.~\eqref{eq:impact} is $\epsilon = 10^{-6}$.

\subsection{Hardware and runtime}
All inference and calibration experiments are run on a single NVIDIA A100 80GB GPU. Compression of one model--sparsity configuration takes between approximately 2{,}000 and 2{,}600 seconds, comparable to ReplaceMe (Cosine) and slower than Streamline; this cost is paid once at compression time and does not affect deployed latency. Inference benchmarks (TTFT and decode speedup in \Cref{tab:inference-speed-kv}) are also run on the same A100 80GB GPU.

\subsection{Reproducibility}
All compression and evaluation runs use a fixed random seed for calibration sampling and token-level sub-sampling. The source code is available at \href{https://github.com/eliacunegatti/SubFit}{https://github.com/eliacunegatti/SubFit}.

\section{Additional Results}
\subsection{Calibration sensitivity}
\label{sec:calibration-sensitivity}

\begin{figure}[!ht]
    \centering
    \includegraphics[width=\linewidth]{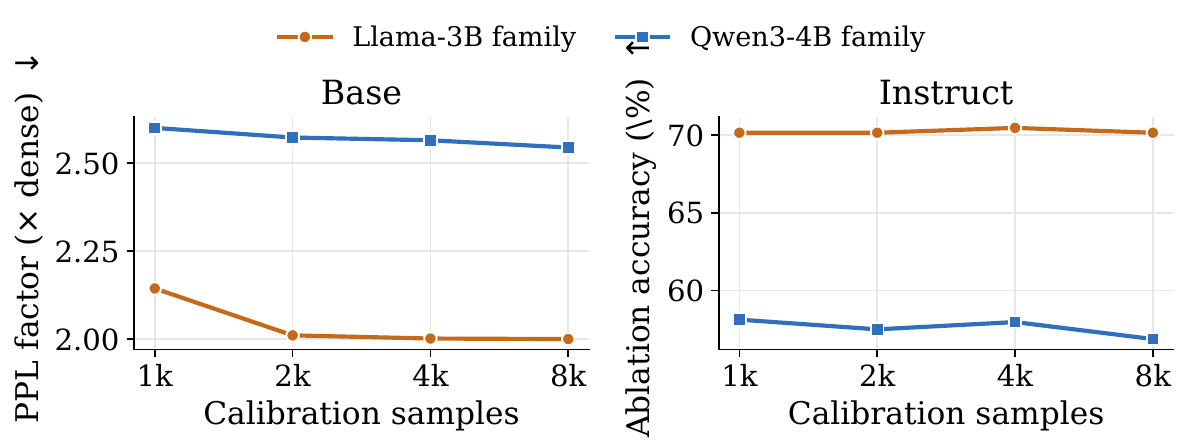}
    \caption{Calibration sample-size sensitivity at $25\%$ sparsity.}
    \label{fig:calibration-sensitivity}
\end{figure}

We analyze the sensitivity of \methodname to both the number of calibration samples and their sequence length. \Cref{fig:calibration-sensitivity} shows that performance is largely stable beyond 2k--4k samples across the evaluated model families. 

\Cref{tab:calibration-sensitivity-values} provides the corresponding task-level results for the Llama-3.2-3B family and the sequence-length sweep. Performance changes only marginally beyond 2k--4k calibration samples, particularly for instruction-tuned accuracy. Increasing sequence length consistently improves base-model perplexity across all three datasets. The selected submodules were identical across the four sequence lengths, suggesting that the improvement does not arise from changes in selection. We hypothesize that longer sequences provide more diverse token statistics and therefore a better-conditioned bypass regression fit. The 8k-sample, 1024-token configuration used in the main experiments represents a practical cost--performance choice rather than evidence that longer calibration contexts provide no benefit.

\begin{table}[!ht]
\label{tab:calibration-sensitivity-values}
\centering
\scriptsize
\setlength{\tabcolsep}{2pt}
\renewcommand{\arraystretch}{1.08}

\begin{tabularx}{\columnwidth}{@{}l c Y Y Y|Y@{}}
\toprule
\multicolumn{6}{c}{(a) Calibration sample-size sensitivity} \\
\midrule
Model & Samples & Lambada $\downarrow$ & C4 $\downarrow$ &
WT2 $\downarrow$ & Inst. Avg. $\uparrow$ \\
\midrule
Base & 1k & 40.14 & 23.64 & 18.10 & -- \\
Base & 2k & 38.71 & 22.94 & 17.15 & -- \\
Base & 4k & 37.85 & 22.70 & 16.90 & -- \\

\rowcolor{lightgreen} Base & 8k & \textbf{37.60} & \textbf{22.58} &
\textbf{16.69} & -- \\
\midrule
Instruct & 1k & -- & -- & -- & 69.53 \\
Instruct & 2k & -- & -- & -- & 70.16 \\ Instruct & 4k & -- & -- & -- & \textbf{70.47} \\
\rowcolor{lightgreen}Instruct & 8k & -- & -- & -- & 70.31 \\
\bottomrule
\end{tabularx}

\vspace{5pt}

\begin{tabularx}{\columnwidth}{@{}Y Y Y|Y@{}}
\toprule
\multicolumn{4}{c}{(b) Calibration sequence-length sensitivity} \\
\midrule
Seq. length & Lambada $\downarrow$ & C4 $\downarrow$ & WT2 $\downarrow$ \\
\midrule
256  & 43.57 & 26.13 & 20.21 \\
512  & 42.53 & 25.22 & 19.06 \\
\rowcolor{lightgreen}
1024 & 37.60 & 22.58 & 16.69 \\
2048 & \textbf{36.72} & \textbf{22.47} & \textbf{16.55} \\
\bottomrule
\end{tabularx}
\caption{Calibration sensitivity on the Llama-3.2-3B family at $25\%$ sparsity. Table (a) varies the number of calibration samples at sequence length 1024. Table (b) varies sequence length using 8k calibration samples. Green rows indicate the configuration used in the main experiments.}
\end{table}

\subsection{Offline runtime}

\Cref{tab:runtime-summary} reports the average runtime of the post-training compression and evaluation pipeline. These values include selection, fitting, and evaluation, and should therefore be interpreted as offline compression cost rather than deployed inference latency. \methodname is slower than simpler pruning baselines in this stage because it performs sequential selection and closed-form residual fitting for both Attention and FFN replacements.

\begin{table}[!ht]
\centering
\label{tab:runtime-summary}
\scriptsize
\setlength{\tabcolsep}{5pt}
\renewcommand{\arraystretch}{1.05}

\begin{tabular}{lccccc}
\toprule
\multicolumn{6}{c}{(a) Base PPL runtime (s) $\downarrow$} \\
\midrule
Method & 12.5\% & 20\% & 25\% & 30\% & 37.5\% \\
\midrule
Streamline (FFN)
    & 764 & \textbf{766} & 912 & 740 & 720 \\
Streamline (Layer)
    & \textbf{683} & 1224 & \textbf{612} & \textbf{723} & \textbf{597} \\
ReplaceMe (LS)
    & 1273 & 1342 & 1141 & 1333 & 1160 \\
ReplaceMe (Cosine)
    & 1662 & 1631 & 1808 & 1643 & 1624 \\
\rowcolor{lightgreen}
\methodname
    & 1905 & 2436 & 2231 & 2673 & 2410 \\
\bottomrule
\end{tabular}

\vspace{4pt}

\begin{tabular}{lccccc}
\toprule
\multicolumn{6}{c}{(b) Instruction-tuned runtime (s) $\downarrow$} \\
\midrule
Method & 12.5\% & 20\% & 25\% & 30\% & 37.5\% \\
\midrule
Streamline (FFN)
    & \textbf{405} & 472 & \textbf{429} & \textbf{390} & \textbf{367} \\
Streamline (Layer)
    & 484 & \textbf{431} & 523 & 407 & 422 \\
ReplaceMe (LS)
    & 1012 & 1021 & 1206 & 1015 & 1013 \\
ReplaceMe (Cosine)
    & 2672 & 2734 & 2692 & 2604 & 2659 \\
\rowcolor{lightgreen}
\methodname
    & 2028 & 2100 & 2137 & 2155 & 2550 \\
\bottomrule
\end{tabular}
\caption{Average pruning runtime in seconds across models for base PPL and instruction-tuned downstream experiments. Lower is better.}
\end{table}

\subsection{Matched inference-efficiency comparison}
The detailed parameter and MAC accounting in \Cref{tab:params-full,tab:macs-full} does not by itself capture realized generation speed. We therefore compare \methodname with the replacement-based baselines under a matched PyTorch generation protocol on Llama-3.2-3B at 25\% sparsity. Unlike the theoretical MAC accounting in \Cref{tab:macs-full}, which uses a sequence length of 2048, this benchmark uses a 512-token prompt.

\begin{table}[!ht]
\label{tab:matched-baseline-efficiency}
\centering
\scriptsize
\setlength{\tabcolsep}{2pt}
\renewcommand{\arraystretch}{1.08}
\begin{tabularx}{\columnwidth}{@{}l Y Y Y Y Y@{}}
\toprule
Method & Params $\downarrow$ & MACs $\downarrow$ &
TTFT $\uparrow$ & Decode $\uparrow$ & KV saved \\
\midrule
Streamline FFN
    & 10.7\% & 14.3\% & \textbf{1.41$\times$}
    & \textbf{1.37$\times$} & 25.0\% \\
Streamline Layer
    & 14.3\% & 14.3\% & 1.38$\times$
    & 1.33$\times$ & 25.0\% \\
ReplaceMe LS
    & \textbf{0.0\%} & \textbf{0.0\%} & 1.37$\times$
    & 1.33$\times$ & 25.0\% \\
ReplaceMe Cosine
    & \textbf{0.0\%} & \textbf{0.0\%} & 1.37$\times$
    & 1.33$\times$ & 25.0\% \\
\rowcolor{lightgreen}
\methodname
    & 12.3\% & 18.1\% & 1.18$\times$
    & 1.12$\times$ & 25.0\% \\
\bottomrule
\end{tabularx}
\caption{Matched efficiency comparison on Llama-3.2-3B at 25\% sparsity using a 512-token prompt and a single seed. Params and MACs report the ratio of added compensation cost to removed cost; speedups are measured relative to the dense model.}
\end{table}

All methods obtain the same KV-cache reduction because they remove the same fraction of Attention computation. However, \methodname achieves smaller TTFT and decode speedups than the baselines because its explicit residual bypasses introduce additional deployed computation. Its improvements in post-compression quality therefore come with an explicit accuracy--efficiency trade-off rather than an unconditional efficiency advantage.

\subsection{GSM8K}
\label{sec:gsm8k}

\Cref{tab:gsm8k-full} reports GSM8K accuracy for all instruction-tuned models and sparsity levels. The results show large variation across methods, especially at low and moderate sparsity, and many compressed models collapse to very low GSM8K accuracy at higher sparsity. This supports our decision to report GSM8K separately rather than include it in the main downstream aggregate (see the limitations discussed in \Cref{sec:limitations}).

\newpage

\begin{table}[!ht]
\label{tab:gsm8k-full}
\centering
\scriptsize
\setlength{\tabcolsep}{2.5pt}
\renewcommand{\arraystretch}{1.04}

\resizebox{\columnwidth}{!}{%
\begin{tabular}{lrrrrr|r}
\toprule
Sparsity. & Streamline (FFN) & Streamline (Layer) & ReplaceMe (LS) & ReplaceMe (Cosine) & \methodname & $\Delta$ \\
\midrule

\multicolumn{7}{c}{\texttt{\textbf{Llama-3.2-3B-Instruct}}} \\
\midrule
\rowcolor{lightgrey} \multicolumn{7}{c}{Dense: 67.7} \\ 
\midrule
12.5\% & 32.1 & \textbf{47.5} & 40.8 & 40.9 & 2.3 & 45.3 \\
20\%   & 2.6  & \textbf{11.6} & 9.1 & 8.5 & 2.7 & 9.0 \\
25\%   & 3.5  & 3.8 & \textbf{5.0} & 3.2 & 1.1 & 3.9 \\
30\%   & 2.0  & 2.4 & \textbf{2.9} & 2.1 & 1.8 & 1.1 \\
37.5\% & 1.7  & 1.9 & 2.0 & \textbf{2.4} & 2.1 & 0.6 \\

\midrule
\multicolumn{7}{c}{\texttt{\textbf{Qwen3-4B-Instruct}}} \\
\midrule
\rowcolor{lightgrey} \multicolumn{7}{c}{Dense: 84.0}  \\
\midrule
12.5\% & 38.5 & 31.7 & 12.7 & 39.7 & \textbf{60.7} & 48.0 \\
20\%   & 4.0  & 2.5 & 2.3 & 5.6 & \textbf{22.0} & 19.7 \\
25\%   & 2.0  & 0.5 & 2.0 & 1.7 & \textbf{2.9} & 2.4 \\
30\%   & 0.1  & 0.0 & 1.8 & \textbf{2.3} & 2.2 & 2.3 \\
37.5\% & 0.0  & 0.0 & 1.1 & 0.5 & \textbf{1.4} & 1.4 \\

\midrule

\multicolumn{7}{c}{\texttt{\textbf{Qwen2.5-7B-Instruct}}} \\ 
\midrule
\rowcolor{lightgrey} \multicolumn{7}{c}{Dense: 81.2} \\ 
\midrule
12.5\% & 7.3 & 16.1 & \textbf{31.9} & 30.0 & 24.1 & 24.6 \\
20\%   & 2.0 & 4.1 & 6.3 & 4.5 & \textbf{7.1} & 5.1 \\
25\%   & 1.8 & 1.7 & 2.4 & 2.6 & \textbf{3.6} & 1.8 \\
30\%   & 1.5 & 1.4 & 2.5 & \textbf{2.6} & 1.6 & 1.2 \\
37.5\% & 1.8 & 1.3 & 1.5 & \textbf{2.2} & 1.4 & 0.9 \\

\midrule
\multicolumn{7}{c}{\texttt{\textbf{DeepSeek-7B-chat}}} \\
\midrule
\rowcolor{lightgrey} \multicolumn{7}{c}{Dense: 51.0} \\
\midrule
12.5\% & 24.6 & 23.1 & 24.2 & 22.0 & \textbf{25.4} & 3.4 \\
20\%   & 3.3 & 9.5 & 9.7 & 5.2 & \textbf{12.8} & 9.5 \\
25\%   & 1.6 & 2.0 & 2.0 & 2.3 & \textbf{3.6} & 2.0 \\
30\%   & 0.2 & 1.1 & 1.7 & 1.4 & \textbf{3.3} & 3.1 \\
37.5\% & 0.0 & 0.5 & 1.3 & \textbf{2.2} & 1.7 & 2.2 \\

\midrule
\multicolumn{7}{c}{\texttt{\textbf{Llama-3.1-8B-Instruct}}}
\\
\midrule
\rowcolor{lightgrey} \multicolumn{7}{c}{Dense: 79.2} \\ 
\midrule
12.5\% & 62.1 & 62.3 & \textbf{65.5} & 65.3 & 56.3 & 9.2 \\
20\%   & 39.7 & 31.6 & 40.6 & \textbf{52.8} & 30.6 & 22.2 \\
25\%   & 4.0 & 3.6 & 10.5 & \textbf{16.7} & 8.3 & 13.1 \\
30\%   & 1.7 & 1.4 & \textbf{2.7} & 2.6 & 1.8 & 1.2 \\
37.5\% & 1.5 & 1.7 & 1.7 & \textbf{2.0} & 2.0 & 0.5 \\

\bottomrule
\end{tabular}%
}
\caption{GSM8K accuracy (\%) across instruction-tuned models and sparsity levels. Dense accuracy is reported below each model name. Spread is the gap between the best and worst sparse methods.}
\end{table}

\clearpage

\subsection{Additional base-model perplexity results}
\label{app:ppl}

\Cref{tab:ppl-appendix1,tab:ppl-appendix2} report the remaining base-model perplexity results at $20\%$, $30\%$, and $37.5\%$ sparsity. These results complement the main language-modeling table, which reports $12.5\%$ and $25\%$. Overall, the additional sparsities show the same trend as the main results: \methodname usually provides the lowest PPL degradation, especially as sparsity increases.
The main exceptions are Qwen3-8B and DeepSeek-7B at some sparsity levels, where the strongest baseline remains competitive or slightly better. This is consistent with the results reported in the main text: \methodname improves the aggregate trade-off, but does not dominate every individual model--sparsity setting.

\begin{table}[!ht]
\label{tab:ppl-appendix1}
\centering
\begin{adjustbox}{max width=\columnwidth}
\setlength{\tabcolsep}{4pt}
\begin{tabular}{l l c c c c}
\toprule
Sparsity & Method & Lambada $\downarrow$ & C4 $\downarrow$ & WT2 $\downarrow$ & PPL Deg. $\downarrow$ \\
\midrule
\multicolumn{6}{c}{\texttt{\textbf{Qwen3-8B}}} \\
\midrule
\rowcolor{lightgrey} Dense & & 25.70 & 15.25 & 9.72 & 1.00$\times$ \\
\midrule
 & Streamline (FFN) & \textbf{45.75} & \textbf{27.01} & \textbf{24.48} & \textbf{1.99$\times$} \\
 & Streamline (Layer) & 61.51 & 41.29 & 57.08 & 3.36$\times$ \\
20\% & ReplaceMe (LS) & 77.52 & 58.55 & 101.88 & 4.95$\times$ \\
 & ReplaceMe (Cosine) & 66.91 & 46.79 & 68.68 & 3.84$\times$ \\
\rowcolor{lightgreen} \cellcolor{white}  & \methodname & 50.25 & 30.69 & 28.01 & 2.25$\times$ \\
\cmidrule(lr){1-6}
 & Streamline (FFN) & 129.13 & 91.62 & 140.29 & 7.58$\times$ \\
 & Streamline (Layer) & 138.22 & 83.89 & 123.24 & 7.21$\times$ \\
30\% & ReplaceMe (LS) & 189.33 & 192.84 & 498.45 & 16.84$\times$ \\
 & ReplaceMe (Cosine) & 85.49 & \textbf{50.95} & \textbf{58.54} & \textbf{4.06$\times$} \\
\rowcolor{lightgreen} \cellcolor{white}  & \methodname & \textbf{78.61} & 52.22 & 64.53 & 4.11$\times$ \\
\cmidrule(lr){1-6}
 & Streamline (FFN) & 444.17 & 387.86 & 737.90 & 32.19$\times$ \\
 & Streamline (Layer) & 392.86 & 288.44 & 497.86 & 24.55$\times$ \\
37.5\% & ReplaceMe (LS) & 168.22 & 118.58 & 261.60 & 11.10$\times$ \\
 & ReplaceMe (Cosine) & \textbf{124.01} & \textbf{82.56} & \textbf{106.55} & \textbf{6.59$\times$} \\
\rowcolor{lightgreen} \cellcolor{white}  & \methodname & 129.88 & 103.97 & 174.30 & 8.52$\times$ \\
\midrule
\multicolumn{6}{c}{\texttt{\textbf{DeepSeek-7B}}} \\
\midrule
\rowcolor{lightgrey} Dense & & 14.36 & 9.75 & 6.85 & 1.00$\times$ \\
\midrule
 & Streamline (FFN) & 25.33 & 20.43 & 17.14 & 2.10$\times$ \\
 & Streamline (Layer) & 25.55 & 20.32 & 17.39 & 2.11$\times$ \\
20\% & ReplaceMe (LS) & 23.42 & 18.87 & 15.59 & 1.93$\times$ \\
 & ReplaceMe (Cosine) & \textbf{21.67} & 18.21 & 13.87 & 1.79$\times$ \\
\rowcolor{lightgreen} \cellcolor{white}  & \methodname & 25.22 & \textbf{15.53} & \textbf{12.38} & \textbf{1.72$\times$} \\
\cmidrule(lr){1-6}
 & Streamline (FFN) & 57.44 & 66.55 & 90.21 & 7.11$\times$ \\
 & Streamline (Layer) & 43.40 & 47.06 & 60.41 & 5.05$\times$ \\
30\% & ReplaceMe (LS) & 40.28 & 41.42 & 44.68 & 4.27$\times$ \\
 & ReplaceMe (Cosine) & 53.75 & 49.66 & 61.55 & 5.55$\times$ \\
\rowcolor{lightgreen} \cellcolor{white}  & \methodname & \textbf{36.51} & \textbf{26.96} & \textbf{27.33} & \textbf{3.04$\times$} \\
\cmidrule(lr){1-6}
 & Streamline (FFN) & 64.56 & 65.22 & 103.33 & 7.68$\times$ \\
 & Streamline (Layer) & \textbf{42.70} & \textbf{39.00} & 58.96 & \textbf{4.68$\times$} \\
37.5\% & ReplaceMe (LS) & 60.16 & 72.76 & 78.37 & 7.10$\times$ \\
 & ReplaceMe (Cosine) & 119.19 & 111.82 & 140.86 & 12.51$\times$ \\
\rowcolor{lightgreen} \cellcolor{white}  & \methodname & 50.90 & 43.52 & \textbf{46.13} & 4.74$\times$ \\
\bottomrule
\end{tabular}
\end{adjustbox}
\caption{Perplexity comparison at representative sparsity levels. Lower PPL is better; PPL Deg. is the geometric mean of sparse/dense PPL ratios.}
\end{table}

\begin{table}[!ht]
\label{tab:ppl-appendix2}
\centering
\begin{adjustbox}{max width=\columnwidth}
\setlength{\tabcolsep}{4pt}
\begin{tabular}{l l c c c c}
\toprule
Sparsity & Method & Lambada $\downarrow$ & C4 $\downarrow$ & WT2 $\downarrow$ & PPL Deg. $\downarrow$ \\
\midrule
\multicolumn{6}{c}{\texttt{\textbf{Llama-3.2-3B}}} \\
\midrule
\rowcolor{lightgrey} Dense & & 20.15 & 11.07 & 7.82 & 1.00$\times$ \\
\midrule
 & Streamline (FFN) & 43.68 & 27.35 & 20.47 & 2.41$\times$ \\
 & Streamline (Layer) & 38.15 & 25.80 & 25.05 & 2.42$\times$ \\
20\% & ReplaceMe (LS) & 42.61 & 27.13 & 27.83 & 2.64$\times$ \\
 & ReplaceMe (Cosine) & 43.74 & 28.39 & 27.37 & 2.69$\times$ \\
\rowcolor{lightgreen} \cellcolor{white}  & \methodname & \textbf{31.94} & \textbf{19.06} & \textbf{13.84} & \textbf{1.69$\times$} \\
\cmidrule(lr){1-6}
 & Streamline (FFN) & 65.94 & 43.51 & 54.44 & 4.47$\times$ \\
 & Streamline (Layer) & 55.31 & 41.39 & 49.86 & 4.03$\times$ \\
30\% & ReplaceMe (LS) & 66.21 & 44.22 & 51.57 & 4.42$\times$ \\
 & ReplaceMe (Cosine) & 67.11 & 45.64 & 50.56 & 4.46$\times$ \\
\rowcolor{lightgreen} \cellcolor{white}  & \methodname & \textbf{49.67} & \textbf{26.23} & \textbf{24.35} & \textbf{2.63$\times$} \\
\cmidrule(lr){1-6}
 & Streamline (FFN) & 129.14 & 86.77 & 134.00 & 9.51$\times$ \\
 & Streamline (Layer) & 108.99 & 74.83 & 105.90 & 7.91$\times$ \\
37.5\% & ReplaceMe (LS) & 94.43 & 64.26 & 94.53 & 6.90$\times$ \\
 & ReplaceMe (Cosine) & 108.02 & 72.42 & 104.68 & 7.77$\times$ \\
\rowcolor{lightgreen} \cellcolor{white}  & \methodname & \textbf{75.27} & \textbf{45.02} & \textbf{58.62} & \textbf{4.85$\times$} \\
\midrule
\multicolumn{6}{c}{\texttt{\textbf{Qwen3-4B}}} \\
\midrule
\rowcolor{lightgrey} Dense & & 33.80 & 19.89 & 13.67 & 1.00$\times$ \\
\midrule
 & Streamline (FFN) & 95.64 & 63.30 & 88.98 & 3.88$\times$ \\
 & Streamline (Layer) & 78.93 & 52.93 & 71.06 & 3.18$\times$ \\
20\% & ReplaceMe (LS) & 94.79 & 61.99 & 83.00 & 3.76$\times$ \\
 & ReplaceMe (Cosine) & 74.98 & 49.95 & 48.77 & 2.71$\times$ \\
\rowcolor{lightgreen} \cellcolor{white}  & \methodname & \textbf{54.36} & \textbf{32.63} & \textbf{30.01} & \textbf{1.80$\times$} \\
\cmidrule(lr){1-6}
 & Streamline (FFN) & 273.13 & 212.23 & 377.12 & 13.35$\times$ \\
 & Streamline (Layer) & 323.62 & 275.06 & 546.83 & 17.43$\times$ \\
30\% & ReplaceMe (LS) & 261.67 & 286.36 & 597.41 & 16.95$\times$ \\
 & ReplaceMe (Cosine) & 440.84 & 522.68 & 642.38 & 25.25$\times$ \\
\rowcolor{lightgreen} \cellcolor{white}  & \methodname & \textbf{92.19} & \textbf{58.99} & \textbf{72.37} & \textbf{3.50$\times$} \\
\cmidrule(lr){1-6}
 & Streamline (FFN) & 721.51 & 726.79 & 1344.13 & 42.49$\times$ \\
 & Streamline (Layer) & 718.56 & 662.92 & 1187.55 & 39.48$\times$ \\
37.5\% & ReplaceMe (LS) & 522.03 & 552.81 & 1206.99 & 33.59$\times$ \\
 & ReplaceMe (Cosine) & 5394.41 & 6397.86 & 10402.30 & 339.31$\times$ \\
\rowcolor{lightgreen} \cellcolor{white}  & \methodname & \textbf{124.51} & \textbf{99.91} & \textbf{140.65} & \textbf{5.75$\times$} \\
\midrule
\multicolumn{6}{c}{\texttt{\textbf{Llama-3.1-8B}}} \\
\midrule
\rowcolor{lightgrey} Dense & & 17.78 & 9.36 & 6.24 & 1.00$\times$ \\
\midrule
 & Streamline (FFN) & 28.55 & 19.58 & 18.39 & 2.15$\times$ \\
 & Streamline (Layer) & \textbf{25.79} & 18.26 & 12.51 & 1.78$\times$ \\
20\% & ReplaceMe (LS) & 28.16 & 18.28 & 13.40 & 1.88$\times$ \\
 & ReplaceMe (Cosine) & 26.59 & 18.29 & 12.68 & 1.81$\times$ \\
\rowcolor{lightgreen} \cellcolor{white}  & \methodname & 28.04 & \textbf{15.60} & \textbf{10.90} & \textbf{1.66$\times$} \\
\cmidrule(lr){1-6}
 & Streamline (FFN) & 58.06 & 44.45 & 65.27 & 5.45$\times$ \\
 & Streamline (Layer) & 49.86 & 44.68 & 56.65 & 4.95$\times$ \\
30\% & ReplaceMe (LS) & 68.58 & 54.84 & 73.81 & 6.44$\times$ \\
 & ReplaceMe (Cosine) & 99.62 & 73.21 & 90.50 & 8.60$\times$ \\
\rowcolor{lightgreen} \cellcolor{white}  & \methodname & \textbf{44.05} & \textbf{27.27} & \textbf{23.40} & \textbf{3.00$\times$} \\
\cmidrule(lr){1-6}
 & Streamline (FFN) & 98.14 & 78.12 & 157.99 & 10.52$\times$ \\
 & Streamline (Layer) & 81.19 & 71.12 & 102.69 & 8.29$\times$ \\
37.5\% & ReplaceMe (LS) & 109.84 & 84.32 & 140.14 & 10.77$\times$ \\
 & ReplaceMe (Cosine) & 346.09 & 152.17 & 317.25 & 25.24$\times$ \\
\rowcolor{lightgreen} \cellcolor{white}  & \methodname & \textbf{69.11} & \textbf{43.69} & \textbf{62.07} & \textbf{5.65$\times$} \\
\bottomrule
\end{tabular}
\end{adjustbox}
\caption{Perplexity comparison at representative sparsity levels. Lower PPL is better; PPL Deg. is the geometric mean of sparse/dense PPL ratios.}
\end{table}

\subsection{Additional instruction-tuned downstream results}
\label{app:instruct}

\Cref{tab:instruct-s0125,tab:instruct-s020,tab:instruct-s030,tab:instruct-s0375} report task-level instruction-tuned results at the remaining sparsity levels. 
These tables provide the per-task breakdown behind the aggregate retention values in
\Cref{tab:summary-retention-ppl}. Across sparsities, \methodname generally preserves the strongest downstream average, while some individual tasks and models remain better served by specific baselines.

\begin{table*}[!t]
\label{tab:instruct-s0125}
\centering
\begin{adjustbox}{max width=\textwidth}
\begin{tabular}{l c c c c c c c c c c c c c | c c}
\toprule
\multicolumn{2}{c}{} & \multicolumn{5}{c}{Commonsense} & \multicolumn{2}{c}{QA} & \multicolumn{1}{c}{Knowledge} & \multicolumn{2}{c}{Reasoning} & \multicolumn{1}{c}{Math} & \multicolumn{1}{c}{Science} & \multicolumn{2}{c}{Average} \\
\cmidrule(lr){3-7} \cmidrule(lr){8-9} \cmidrule(lr){10-10} \cmidrule(lr){11-12} \cmidrule(lr){13-13} \cmidrule(lr){14-14} \cmidrule(lr){15-16}
Model & Sparsity & HS & PIQA & Wino & OBQA & SIQA & BoolQ & TQA & MMLU & ARC-C & ARC-E & MathQA & SciQ & Avg & Acc. Ret. \\
\midrule

\rowcolor{lightgrey}\texttt{\textbf{Llama-3.2-3B-Instruct}} & Dense & 64.53 & 75.63 & 67.88 & 42.00 & 46.11 & 75.63 & 33.54 & 61.84 & 45.31 & 68.10 & 26.03 & 87.90 & 57.87 & 100.00\% \\
Streamline (FFN) &  & 59.91 & \textbf{71.16} & 67.40 & 35.00 & \textbf{45.80} & 74.19 & \textbf{31.58} & 50.03 & 39.42 & 59.68 & 23.82 & 83.40 & 53.45 & 92.36\% \\
Streamline (Layer) &  & \textbf{60.16} & 70.40 & 65.90 & 35.00 & 45.29 & 75.29 & 31.46 & 56.20 & 38.57 & 59.81 & \textbf{24.46} & 86.00 & 54.04 & 93.38\% \\
ReplaceMe (LS) & 12.5\% & 58.32 & 70.40 & \textbf{68.51} & \textbf{36.40} & 44.73 & \textbf{80.18} & 30.60 & \textbf{61.65} & 37.88 & 57.66 & 23.79 & 81.70 & 54.32 & 93.86\% \\
ReplaceMe (Cosine) &  & 58.94 & 70.40 & \textbf{68.51} & 36.00 & 44.58 & 79.57 & 30.48 & 61.51 & 38.65 & 58.80 & 23.85 & 83.60 & 54.57 & 94.30\% \\
\rowcolor{lightgreen}\methodname &  & 58.88 & 69.15 & 68.11 & 35.20 & 45.24 & 79.88 & 29.99 & 61.07 & \textbf{39.51} & \textbf{60.23} & 23.08 & \textbf{89.40} & \textbf{54.98} & \textbf{95.00\%} \\

\midrule

\rowcolor{lightgrey}\texttt{\textbf{Qwen3-4B-Instruct}} & Dense & 43.42 & 69.70 & 56.75 & 40.80 & 46.32 & 84.83 & 39.17 & 72.18 & 43.17 & 55.43 & 33.30 & 67.50 & 54.38 & 100.00\% \\
Streamline (FFN) &  & 39.39 & 63.22 & 58.48 & 34.80 & 40.89 & 83.52 & 38.80 & 71.24 & 36.95 & 43.14 & 28.27 & 62.90 & 50.13 & 92.19\% \\
Streamline (Layer) &  & 40.50 & 65.56 & \textbf{59.43} & 36.60 & 42.58 & 83.61 & \textbf{39.78} & 71.41 & 37.88 & 48.19 & 28.04 & 63.40 & \textbf{51.42} & \textbf{94.55\%} \\
ReplaceMe (LS) & 12.5\% & \textbf{44.49} & \textbf{70.46} & 56.04 & 39.00 & 43.96 & 73.46 & 31.70 & 47.48 & \textbf{41.72} & \textbf{56.06} & 27.47 & \textbf{72.30} & 50.34 & 92.58\% \\
ReplaceMe (Cosine) &  & 44.38 & 70.24 & 54.46 & \textbf{39.80} & \textbf{44.32} & 75.87 & 32.68 & 49.57 & 40.36 & 55.60 & 28.48 & 68.50 & 50.35 & 92.60\% \\
\rowcolor{lightgreen}\methodname &  & 36.05 & 62.24 & 58.17 & 37.20 & 43.14 & \textbf{85.72} & \textbf{39.78} & \textbf{72.08} & 34.73 & 44.65 & \textbf{29.08} & 63.70 & 50.55 & 92.95\% \\

\midrule

\rowcolor{lightgrey}\texttt{\textbf{Qwen2.5-7B-Instruct}} & Dense & 65.48 & 73.50 & 61.01 & 44.00 & 45.75 & 85.81 & 47.49 & 73.51 & 43.34 & 50.84 & 34.17 & 55.40 & 56.69 & 100.00\% \\
Streamline (FFN) &  & \textbf{60.88} & \textbf{75.14} & 59.04 & 39.40 & 42.58 & 71.65 & 28.03 & 40.08 & 38.40 & 44.15 & 30.28 & \textbf{55.20} & 48.74 & 85.97\% \\
Streamline (Layer) &  & 60.57 & 74.81 & 61.64 & 42.20 & 43.04 & 75.11 & 28.15 & 47.42 & 38.91 & 46.51 & 30.89 & 54.70 & 50.33 & 88.77\% \\
ReplaceMe (LS) & 12.5\% & 60.26 & 71.87 & \textbf{62.43} & \textbf{43.00} & 44.47 & 75.54 & 32.44 & 49.41 & 39.42 & \textbf{47.26} & 30.92 & 51.40 & \textbf{50.70} & \textbf{89.43\%} \\
ReplaceMe (Cosine) &  & 58.93 & 72.96 & 61.80 & 42.40 & \textbf{44.78} & 72.42 & 31.58 & 49.42 & \textbf{40.10} & 46.76 & \textbf{31.22} & 51.80 & 50.35 & 88.81\% \\
\rowcolor{lightgreen}\methodname &  & 56.11 & 72.96 & 58.01 & 41.60 & 43.09 & \textbf{81.01} & \textbf{34.76} & \textbf{51.05} & 36.86 & 46.13 & 29.82 & 49.60 & 50.08 & 88.34\% \\

\midrule

\rowcolor{lightgrey}\texttt{\textbf{DeepSeek-7B-chat}} & Dense & 70.56 & 77.64 & 74.90 & 44.80 & 50.46 & 83.98 & 37.21 & 50.90 & 41.55 & 62.63 & 27.54 & 78.40 & 58.38 & 100.00\% \\
Streamline (FFN) &  & 63.66 & 73.83 & 72.69 & 39.80 & 47.39 & 83.12 & 32.93 & 51.05 & 38.91 & 55.30 & 25.36 & 67.40 & 54.29 & 92.99\% \\
Streamline (Layer) &  & 63.35 & 73.61 & 73.24 & 39.80 & 47.39 & 84.04 & \textbf{33.90} & 51.13 & 37.80 & 56.02 & 26.87 & 70.80 & 54.83 & 93.92\% \\
ReplaceMe (LS) & 12.5\% & 61.79 & 73.61 & 73.80 & 40.00 & \textbf{47.95} & 83.70 & 33.41 & 50.80 & 37.88 & 54.29 & 27.00 & 67.20 & 54.29 & 92.99\% \\
ReplaceMe (Cosine) &  & \textbf{63.87} & \textbf{73.88} & \textbf{73.95} & 41.20 & 47.59 & 80.73 & 32.80 & 50.73 & \textbf{39.68} & \textbf{58.71} & 26.63 & \textbf{77.40} & \textbf{55.60} & \textbf{95.24\%} \\
\rowcolor{lightgreen}\methodname &  & 62.89 & 73.45 & 73.80 & \textbf{41.40} & \textbf{47.95} & \textbf{84.37} & 33.54 & \textbf{51.32} & 38.91 & 56.69 & \textbf{27.27} & 73.50 & 55.42 & 94.94\% \\

\midrule

\rowcolor{lightgrey}\texttt{\textbf{Llama-3.1-8B-Instruct}} & Dense & 72.52 & 79.49 & 77.82 & 49.00 & 50.20 & 83.88 & 40.39 & 68.79 & 53.58 & 75.76 & 26.80 & 91.70 & 64.16 & 100.00\% \\
Streamline (FFN) &  & \textbf{70.12} & 76.71 & \textbf{77.11} & 44.80 & \textbf{49.39} & 73.33 & 36.35 & 67.76 & 49.06 & 72.90 & 25.09 & \textbf{90.60} & 61.10 & 95.23\% \\
Streamline (Layer) &  & 69.60 & 76.88 & \textbf{77.11} & \textbf{45.80} & 49.23 & 70.73 & 37.21 & 67.90 & 48.98 & \textbf{73.70} & 26.03 & 90.50 & 61.14 & 95.29\% \\
ReplaceMe (LS) & 12.5\% & 68.24 & \textbf{77.04} & 76.40 & 45.20 & 48.67 & 73.03 & 37.45 & 68.16 & 48.38 & 72.64 & 25.70 & 88.50 & 60.78 & 94.74\% \\
ReplaceMe (Cosine) &  & 68.70 & 76.55 & 76.32 & 45.20 & 49.33 & 72.32 & \textbf{38.43} & \textbf{68.29} & \textbf{49.32} & 73.02 & 25.49 & 89.50 & 61.04 & 95.14\% \\
\rowcolor{lightgreen}\methodname &  & 68.19 & 76.39 & 75.37 & 45.40 & 49.33 & \textbf{80.89} & 37.45 & 66.65 & 48.63 & 72.69 & \textbf{26.10} & 89.70 & \textbf{61.40} & \textbf{95.70\%} \\

\bottomrule
\end{tabular}
\end{adjustbox}
\caption{Downstream performance (\%) at sparsity 12.5\%. Acc. Ret. is the ratio sparse/dense aggregate accuracy.}
\end{table*}

\begin{table*}[t]
\label{tab:instruct-s020}
\centering
\begin{adjustbox}{max width=\textwidth}
\begin{tabular}{l c c c c c c c c c c c c c | c c}
\toprule
\multicolumn{2}{c}{} & \multicolumn{5}{c}{Commonsense} & \multicolumn{2}{c}{QA} & \multicolumn{1}{c}{Knowledge} & \multicolumn{2}{c}{Reasoning} & \multicolumn{1}{c}{Math} & \multicolumn{1}{c}{Science} & \multicolumn{2}{c}{Average} \\
\cmidrule(lr){3-7} \cmidrule(lr){8-9} \cmidrule(lr){10-10} \cmidrule(lr){11-12} \cmidrule(lr){13-13} \cmidrule(lr){14-14} \cmidrule(lr){15-16}
Model & Sparsity & HS & PIQA & Wino & OBQA & SIQA & BoolQ & TQA & MMLU & ARC-C & ARC-E & MathQA & SciQ & Avg & Acc. Ret. \\
\midrule

\rowcolor{lightgrey}\texttt{\textbf{Llama-3.2-3B-Instruct}} & Dense & 64.53 & 75.63 & 67.88 & 42.00 & 46.11 & 75.63 & 33.54 & 61.84 & 45.31 & 68.10 & 26.03 & 87.90 & 57.87 & 100.00\% \\
Streamline (FFN) &  & 53.44 & 66.87 & \textbf{67.96} & 32.00 & 42.07 & 80.12 & 28.40 & 60.02 & 34.39 & 51.30 & \textbf{23.69} & 72.00 & 51.02 & 88.16\% \\
Streamline (Layer) &  & 54.83 & 66.05 & 66.30 & 32.80 & 42.94 & 62.69 & 30.72 & 35.39 & 36.43 & 53.24 & 23.35 & 74.90 & 48.30 & 83.46\% \\
ReplaceMe (LS) & 20\% & 51.97 & 66.27 & 66.22 & 32.80 & 42.73 & 80.92 & 30.97 & 61.28 & 33.87 & 50.08 & 23.05 & 72.60 & 51.06 & 88.23\% \\
ReplaceMe (Cosine) &  & 53.30 & 66.38 & 67.40 & 33.20 & 42.02 & \textbf{81.71} & 30.97 & \textbf{61.47} & 33.87 & 50.25 & 22.88 & 73.40 & 51.40 & 88.82\% \\
\rowcolor{lightgreen}\methodname &  & \textbf{55.03} & \textbf{67.85} & 65.35 & \textbf{35.20} & \textbf{44.32} & 71.77 & \textbf{31.21} & 61.00 & \textbf{37.46} & \textbf{56.48} & 22.68 & \textbf{81.70} & \textbf{52.50} & \textbf{90.72\%} \\

\midrule

\rowcolor{lightgrey}\texttt{\textbf{Qwen3-4B-Instruct}} & Dense & 43.42 & 69.70 & 56.75 & 40.80 & 46.32 & 84.83 & 39.17 & 72.18 & 43.17 & 55.43 & 33.30 & 67.50 & 54.38 & 100.00\% \\
Streamline (FFN) &  & 32.02 & 54.62 & 55.64 & 31.40 & 37.72 & \textbf{83.64} & 32.19 & \textbf{71.51} & 29.18 & 35.35 & 25.93 & 57.70 & 45.58 & 83.81\% \\
Streamline (Layer) &  & 32.51 & 56.64 & \textbf{57.22} & 33.00 & 38.69 & 73.58 & 34.64 & 71.02 & 30.46 & 37.84 & \textbf{26.73} & 55.70 & 45.67 & 83.98\% \\
ReplaceMe (LS) & 20\% & \textbf{40.13} & 68.17 & 52.96 & \textbf{36.40} & 40.07 & 57.92 & 31.21 & 32.59 & \textbf{35.07} & \textbf{45.75} & 24.99 & 55.00 & 43.36 & 79.73\% \\
ReplaceMe (Cosine) &  & 40.07 & \textbf{68.72} & 53.59 & 35.40 & \textbf{41.66} & 62.54 & 30.60 & 34.57 & 34.47 & 45.37 & 26.13 & 52.90 & 43.83 & 80.61\% \\
\rowcolor{lightgreen}\methodname &  & 30.69 & 56.64 & 55.80 & 34.20 & 41.50 & 72.20 & \textbf{36.72} & 70.51 & 29.95 & 38.30 & \textbf{26.73} & \textbf{62.30} & \textbf{46.30} & \textbf{85.13\%} \\

\midrule

\rowcolor{lightgrey}\texttt{\textbf{Qwen2.5-7B-Instruct}} & Dense & 65.48 & 73.50 & 61.01 & 44.00 & 45.75 & 85.81 & 47.49 & 73.51 & 43.34 & 50.84 & 34.17 & 55.40 & 56.69 & 100.00\% \\
Streamline (FFN) &  & 55.27 & 71.76 & 55.80 & 39.80 & 41.30 & 60.49 & 26.07 & 30.16 & 33.87 & 42.21 & 29.15 & 52.60 & 44.87 & 79.15\% \\
Streamline (Layer) &  & \textbf{56.51} & 73.18 & 55.25 & 40.80 & 42.17 & 65.84 & 26.19 & 31.67 & 34.30 & 44.78 & 28.54 & 54.30 & 46.13 & 81.37\% \\
ReplaceMe (LS) & 20\% & 55.88 & 72.80 & 55.17 & 39.60 & 42.27 & 66.70 & 29.38 & 32.59 & \textbf{37.88} & 45.41 & 29.31 & \textbf{54.50} & 46.79 & 82.53\% \\
ReplaceMe (Cosine) &  & 56.34 & \textbf{73.23} & 57.38 & \textbf{42.40} & \textbf{43.65} & 65.87 & 29.50 & 34.03 & 37.03 & \textbf{45.96} & \textbf{29.75} & 53.90 & \textbf{47.42} & \textbf{83.65\%} \\
\rowcolor{lightgreen}\methodname &  & 51.49 & 71.38 & \textbf{57.70} & 40.00 & 41.04 & \textbf{69.45} & \textbf{33.41} & \textbf{39.45} & 35.07 & 44.40 & 28.64 & 52.30 & 47.03 & 82.95\% \\

\midrule

\rowcolor{lightgrey}\texttt{\textbf{DeepSeek-7B-chat}} & Dense & 70.56 & 77.64 & 74.90 & 44.80 & 50.46 & 83.98 & 37.21 & 50.90 & 41.55 & 62.63 & 27.54 & 78.40 & 58.38 & 100.00\% \\
Streamline (FFN) &  & \textbf{60.70} & 70.40 & 74.35 & 35.60 & 46.01 & 77.49 & 31.82 & 48.40 & 36.18 & 50.34 & 22.75 & 62.90 & 51.41 & 88.06\% \\
Streamline (Layer) &  & 56.61 & 68.50 & 73.64 & 35.80 & 46.11 & 78.65 & \textbf{32.44} & 49.88 & 34.56 & 49.41 & 22.61 & 61.50 & 50.81 & 87.03\% \\
ReplaceMe (LS) & 20\% & 55.15 & 69.04 & \textbf{74.43} & 37.60 & 46.32 & 78.93 & 31.82 & 50.48 & 34.04 & 49.83 & 23.85 & 59.70 & 50.93 & 87.24\% \\
ReplaceMe (Cosine) &  & 59.55 & \textbf{70.57} & 73.09 & 35.60 & 46.37 & 79.54 & \textbf{32.44} & 46.76 & \textbf{36.69} & \textbf{53.24} & 24.59 & \textbf{73.60} & \textbf{52.67} & \textbf{90.22\%} \\
\rowcolor{lightgreen}\methodname &  & 55.21 & 69.80 & 71.03 & \textbf{40.00} & \textbf{46.98} & \textbf{84.28} & \textbf{32.44} & \textbf{50.75} & 35.75 & 52.99 & \textbf{24.82} & 67.20 & 52.60 & 90.11\% \\

\midrule

\rowcolor{lightgrey}\texttt{\textbf{Llama-3.1-8B-Instruct}} & Dense & 72.52 & 79.49 & 77.82 & 49.00 & 50.20 & 83.88 & 40.39 & 68.79 & 53.58 & 75.76 & 26.80 & 91.70 & 64.16 & 100.00\% \\
Streamline (FFN) &  & \textbf{66.37} & \textbf{74.43} & 75.22 & 40.20 & 47.59 & 80.58 & 35.62 & 67.60 & 43.17 & \textbf{65.82} & 23.38 & 85.10 & \textbf{58.76} & \textbf{91.58\%} \\
Streamline (Layer) &  & 65.82 & 74.27 & 74.82 & 39.60 & \textbf{48.16} & 77.31 & 35.99 & 67.72 & \textbf{43.52} & 65.45 & 23.42 & \textbf{86.90} & 58.58 & 91.30\% \\
ReplaceMe (LS) & 20\% & 63.56 & 72.42 & 74.82 & 42.00 & 45.91 & 81.47 & 36.60 & \textbf{68.24} & 43.26 & 65.74 & \textbf{25.03} & 85.10 & 58.68 & 91.45\% \\
ReplaceMe (Cosine) &  & 65.71 & 73.67 & 74.51 & 41.00 & 47.29 & 75.66 & \textbf{37.33} & 68.02 & 42.58 & 64.90 & 24.05 & 83.70 & 58.20 & 90.71\% \\
\rowcolor{lightgreen}\methodname &  & 63.41 & 72.09 & \textbf{75.61} & \textbf{42.20} & 46.93 & \textbf{83.33} & 34.76 & 67.59 & 41.72 & 65.61 & 24.56 & 86.20 & 58.67 & 91.44\% \\

\bottomrule
\end{tabular}
\end{adjustbox}
\caption{Downstream performance (\%) at sparsity 20\%. Acc. Ret. is the ratio sparse/dense aggregate accuracy.}
\end{table*}

\begin{table*}[t]
\label{tab:instruct-s030}
\centering
\begin{adjustbox}{max width=\textwidth}
\begin{tabular}{l c c c c c c c c c c c c c | c c}
\toprule
\multicolumn{2}{c}{} & \multicolumn{5}{c}{Commonsense} & \multicolumn{2}{c}{QA} & \multicolumn{1}{c}{Knowledge} & \multicolumn{2}{c}{Reasoning} & \multicolumn{1}{c}{Math} & \multicolumn{1}{c}{Science} & \multicolumn{2}{c}{Average} \\
\cmidrule(lr){3-7} \cmidrule(lr){8-9} \cmidrule(lr){10-10} \cmidrule(lr){11-12} \cmidrule(lr){13-13} \cmidrule(lr){14-14} \cmidrule(lr){15-16}
Model & Sparsity & HS & PIQA & Wino & OBQA & SIQA & BoolQ & TQA & MMLU & ARC-C & ARC-E & MathQA & SciQ & Avg & Acc. Ret. \\
\midrule

\rowcolor{lightgrey}\texttt{\textbf{Llama-3.2-3B-Instruct}} & Dense & 64.53 & 75.63 & 67.88 & 42.00 & 46.11 & 75.63 & 33.54 & 61.84 & 45.31 & 68.10 & 26.03 & 87.90 & 57.87 & 100.00\% \\
Streamline (FFN) &  & 46.77 & 64.04 & 65.19 & 32.60 & 40.48 & 78.56 & 28.27 & 41.51 & 32.51 & 43.69 & 21.07 & 65.70 & 46.70 & 80.69\% \\
Streamline (Layer) &  & 48.55 & \textbf{64.25} & \textbf{65.51} & 32.40 & 41.20 & 77.46 & 27.66 & 41.60 & \textbf{33.53} & 46.00 & 21.14 & 69.60 & 47.41 & 81.92\% \\
ReplaceMe (LS) & 30\% & 47.74 & 63.06 & 60.93 & 33.80 & 40.48 & \textbf{79.33} & 28.27 & \textbf{60.36} & 31.06 & 46.68 & 22.35 & 68.40 & 48.54 & 83.87\% \\
ReplaceMe (Cosine) &  & \textbf{50.17} & 63.77 & 60.85 & \textbf{34.00} & 40.89 & 76.18 & 29.99 & 56.16 & 32.08 & \textbf{47.73} & \textbf{23.35} & 70.60 & 48.81 & 84.34\% \\
\rowcolor{lightgreen}\methodname &  & 48.51 & 63.44 & 64.64 & 32.00 & \textbf{41.61} & 78.99 & \textbf{30.60} & 60.16 & 32.94 & 47.14 & 21.84 & \textbf{75.70} & \textbf{49.80} & \textbf{86.04\%} \\

\midrule

\rowcolor{lightgrey}\texttt{\textbf{Qwen3-4B-Instruct}} & Dense & 43.42 & 69.70 & 56.75 & 40.80 & 46.32 & 84.83 & 39.17 & 72.18 & 43.17 & 55.43 & 33.30 & 67.50 & 54.38 & 100.00\% \\
Streamline (FFN) &  & 23.69 & 51.58 & 51.62 & 32.40 & 34.85 & \textbf{68.87} & 24.85 & 22.94 & 25.60 & 30.13 & 20.47 & 21.80 & 34.07 & 62.64\% \\
Streamline (Layer) &  & 22.51 & 48.53 & 51.38 & 33.60 & 34.44 & 63.52 & 25.70 & 27.29 & 25.00 & 28.16 & 21.81 & 20.60 & 33.54 & 61.69\% \\
ReplaceMe (LS) & 30\% & 33.94 & \textbf{63.82} & 53.04 & \textbf{34.20} & 36.85 & 44.92 & 37.70 & 26.42 & \textbf{29.44} & \textbf{39.27} & \textbf{24.36} & 51.90 & 39.65 & 72.92\% \\
ReplaceMe (Cosine) &  & \textbf{35.06} & 62.02 & 50.99 & 33.60 & 38.02 & 56.73 & \textbf{38.07} & 26.93 & 29.10 & 36.03 & 23.48 & 53.70 & 40.31 & 74.13\% \\
\rowcolor{lightgreen}\methodname &  & 28.71 & 54.57 & \textbf{55.25} & 33.60 & \textbf{39.10} & 65.20 & 33.41 & \textbf{56.30} & 27.39 & 35.10 & 24.19 & \textbf{54.80} & \textbf{42.30} & \textbf{77.79\%} \\

\midrule

\rowcolor{lightgrey}\texttt{\textbf{Qwen2.5-7B-Instruct}} & Dense & 65.48 & 73.50 & 61.01 & 44.00 & 45.75 & 85.81 & 47.49 & 73.51 & 43.34 & 50.84 & 34.17 & 55.40 & 56.69 & 100.00\% \\
Streamline (FFN) &  & 49.67 & 71.55 & 51.78 & 39.40 & 40.12 & 62.17 & 24.11 & 26.79 & 32.25 & 43.01 & 25.90 & \textbf{57.80} & 43.71 & 77.11\% \\
Streamline (Layer) &  & 50.49 & \textbf{72.09} & 53.28 & 38.60 & 39.92 & 62.17 & 23.99 & 26.73 & 32.85 & 43.10 & 26.03 & 55.90 & 43.76 & 77.19\% \\
ReplaceMe (LS) & 30\% & 50.50 & 68.99 & 56.20 & 39.40 & 42.43 & 62.14 & 29.99 & 28.98 & 34.13 & 42.00 & \textbf{27.77} & 49.90 & 44.37 & 78.26\% \\
ReplaceMe (Cosine) &  & \textbf{50.82} & 70.19 & \textbf{58.64} & \textbf{40.60} & \textbf{42.94} & 63.27 & 30.35 & 29.24 & \textbf{34.22} & \textbf{44.07} & 27.64 & 55.20 & \textbf{45.60} & \textbf{80.43\%} \\
\rowcolor{lightgreen}\methodname &  & 45.65 & 68.77 & 54.14 & 39.00 & 40.58 & \textbf{66.57} & \textbf{30.97} & \textbf{33.13} & 32.51 & 41.54 & 26.00 & 51.00 & 44.16 & 77.89\% \\

\midrule

\rowcolor{lightgrey}\texttt{\textbf{DeepSeek-7B-chat}} & Dense & 70.56 & 77.64 & 74.90 & 44.80 & 50.46 & 83.98 & 37.21 & 50.90 & 41.55 & 62.63 & 27.54 & 78.40 & 58.38 & 100.00\% \\
Streamline (FFN) &  & 26.71 & 54.73 & 61.56 & 28.80 & 36.49 & 63.36 & 27.05 & 29.50 & 29.69 & 32.74 & 18.86 & 30.60 & 36.68 & 62.82\% \\
Streamline (Layer) &  & 46.52 & 64.15 & 69.22 & \textbf{34.80} & 40.43 & 63.27 & 25.21 & \textbf{45.25} & 28.84 & 39.52 & 22.85 & 42.90 & 43.58 & 74.65\% \\
ReplaceMe (LS) & 30\% & 42.85 & 60.28 & 68.11 & 33.60 & 40.69 & 67.09 & 26.81 & 40.31 & 29.01 & 37.92 & 22.38 & 34.60 & 41.97 & 71.89\% \\
ReplaceMe (Cosine) &  & \textbf{47.81} & \textbf{70.84} & 53.20 & 33.80 & 40.43 & 53.52 & 26.68 & 25.59 & 26.37 & \textbf{47.98} & 23.58 & \textbf{73.90} & 43.64 & 74.75\% \\
\rowcolor{lightgreen}\methodname &  & 44.78 & 62.24 & \textbf{70.01} & 31.80 & \textbf{43.76} & \textbf{79.45} & \textbf{28.15} & 37.69 & \textbf{29.95} & 41.92 & \textbf{23.99} & 49.40 & \textbf{45.26} & \textbf{77.53\%} \\

\midrule

\rowcolor{lightgrey}\texttt{\textbf{Llama-3.1-8B-Instruct}} & Dense & 72.52 & 79.49 & 77.82 & 49.00 & 50.20 & 83.88 & 40.39 & 68.79 & 53.58 & 75.76 & 26.80 & 91.70 & 64.16 & 100.00\% \\
Streamline (FFN) &  & 49.18 & 62.79 & 72.14 & 32.00 & 41.25 & 82.35 & 32.80 & 62.19 & 33.53 & 44.32 & 21.04 & 69.60 & 50.27 & 78.34\% \\
Streamline (Layer) &  & 49.41 & 63.00 & 71.35 & 32.60 & 41.81 & 81.90 & 31.21 & 62.28 & 35.49 & 47.10 & 21.68 & 70.40 & 50.69 & 79.00\% \\
ReplaceMe (LS) & 30\% & 50.92 & 64.09 & 71.11 & 34.80 & 41.35 & 84.04 & 33.41 & \textbf{67.18} & 33.62 & 49.71 & 21.54 & 75.10 & 52.24 & 81.42\% \\
ReplaceMe (Cosine) &  & 53.18 & 65.94 & 71.51 & \textbf{36.20} & 42.68 & 84.22 & \textbf{36.60} & 66.42 & 34.47 & 46.55 & 22.08 & 71.90 & 52.64 & 82.05\% \\
\rowcolor{lightgreen}\methodname &  & \textbf{54.26} & \textbf{67.14} & \textbf{72.53} & 35.40 & \textbf{43.40} & \textbf{84.25} & 34.27 & 66.63 & \textbf{36.26} & \textbf{53.20} & \textbf{22.88} & \textbf{82.20} & \textbf{54.37} & \textbf{84.74\%} \\

\bottomrule
\end{tabular}
\end{adjustbox}
\caption{Downstream performance (\%) at sparsity 30\%. Acc. Ret. is the ratio sparse/dense aggregate accuracy.}
\end{table*}

\begin{table*}[t]
\label{tab:instruct-s0375}
\centering
\begin{adjustbox}{max width=\textwidth}
\begin{tabular}{l c c c c c c c c c c c c c | c c}
\toprule
\multicolumn{2}{c}{} & \multicolumn{5}{c}{Commonsense} & \multicolumn{2}{c}{QA} & \multicolumn{1}{c}{Knowledge} & \multicolumn{2}{c}{Reasoning} & \multicolumn{1}{c}{Math} & \multicolumn{1}{c}{Science} & \multicolumn{2}{c}{Average} \\
\cmidrule(lr){3-7} \cmidrule(lr){8-9} \cmidrule(lr){10-10} \cmidrule(lr){11-12} \cmidrule(lr){13-13} \cmidrule(lr){14-14} \cmidrule(lr){15-16}
Model & Sparsity & HS & PIQA & Wino & OBQA & SIQA & BoolQ & TQA & MMLU & ARC-C & ARC-E & MathQA & SciQ & Avg & Acc. Ret. \\
\midrule

\rowcolor{lightgrey}\texttt{\textbf{Llama-3.2-3B-Instruct}} & Dense & 64.53 & 75.63 & 67.88 & 42.00 & 46.11 & 75.63 & 33.54 & 61.84 & 45.31 & 68.10 & 26.03 & 87.90 & 57.87 & 100.00\% \\
Streamline (FFN) &  & 42.20 & 60.39 & \textbf{64.48} & 31.00 & 38.79 & \textbf{76.64} & 27.42 & 48.68 & 31.23 & 38.64 & 20.34 & 29.20 & 42.42 & 73.29\% \\
Streamline (Layer) &  & \textbf{45.90} & \textbf{61.70} & 64.09 & \textbf{31.40} & 39.61 & 75.99 & 27.17 & \textbf{53.78} & 31.23 & 39.77 & 20.94 & 47.10 & 44.89 & 77.57\% \\
ReplaceMe (LS) & 37.5\% & 41.39 & 59.96 & 64.33 & 29.80 & 38.69 & 73.21 & 26.81 & 52.97 & 30.03 & 40.61 & 21.47 & 60.00 & 44.94 & 77.65\% \\
ReplaceMe (Cosine) &  & 45.09 & 61.32 & 62.19 & 29.60 & \textbf{39.66} & 67.49 & \textbf{29.62} & 25.35 & \textbf{31.40} & 40.95 & \textbf{22.61} & 68.70 & 43.67 & 75.45\% \\
\rowcolor{lightgreen}\methodname &  & 43.09 & 60.56 & 64.25 & 30.60 & 39.51 & 74.37 & \textbf{29.62} & 36.00 & 30.72 & \textbf{42.85} & 22.35 & \textbf{69.50} & \textbf{45.28} & \textbf{78.24\%} \\

\midrule

\rowcolor{lightgrey}\texttt{\textbf{Qwen3-4B-Instruct}} & Dense & 43.42 & 69.70 & 56.75 & 40.80 & 46.32 & 84.83 & 39.17 & 72.18 & 43.17 & 55.43 & 33.30 & 67.50 & 54.38 & 100.00\% \\
Streamline (FFN) &  & 25.45 & 51.69 & \textbf{52.49} & \textbf{36.00} & 34.03 & 37.83 & 22.77 & 22.92 & \textbf{28.07} & 29.12 & 19.40 & 23.60 & 31.95 & 58.75\% \\
Streamline (Layer) &  & 20.87 & 46.52 & 49.09 & 32.60 & 33.62 & 64.28 & 23.01 & 22.95 & 24.83 & 25.59 & 20.07 & 18.10 & 31.79 & 58.47\% \\
ReplaceMe (LS) & 37.5\% & 20.21 & 48.64 & 49.96 & 30.80 & 34.14 & 67.22 & 25.46 & 23.34 & 23.89 & 30.56 & 21.47 & 23.10 & 33.23 & 61.11\% \\
ReplaceMe (Cosine) &  & 23.81 & 50.98 & 52.25 & 30.00 & 33.06 & 46.42 & 24.72 & 24.01 & 26.54 & 32.41 & 18.79 & 30.60 & 32.80 & 60.31\% \\
\rowcolor{lightgreen}\methodname &  & \textbf{28.08} & \textbf{54.90} & 51.70 & 32.00 & \textbf{37.92} & \textbf{71.01} & \textbf{31.33} & \textbf{41.01} & 26.19 & \textbf{33.04} & \textbf{24.09} & \textbf{52.50} & \textbf{40.31} & \textbf{74.13\%} \\

\midrule

\rowcolor{lightgrey}\texttt{\textbf{Qwen2.5-7B-Instruct}} & Dense & 65.48 & 73.50 & 61.01 & 44.00 & 45.75 & 85.81 & 47.49 & 73.51 & 43.34 & 50.84 & 34.17 & 55.40 & 56.69 & 100.00\% \\
Streamline (FFN) &  & 41.98 & 64.85 & 52.09 & 35.00 & 38.43 & 44.89 & 30.11 & 25.26 & 27.73 & 38.09 & 25.53 & 51.10 & 39.59 & 69.83\% \\
Streamline (Layer) &  & \textbf{45.17} & \textbf{67.03} & 51.38 & 37.20 & 39.10 & 39.08 & 30.84 & 26.09 & 29.69 & 37.46 & 26.30 & 50.80 & 40.01 & 70.58\% \\
ReplaceMe (LS) & 37.5\% & 43.91 & 66.05 & \textbf{53.67} & 36.20 & 39.15 & 61.74 & 29.13 & \textbf{28.17} & \textbf{31.06} & 40.24 & 26.53 & 51.80 & 42.30 & 74.62\% \\
ReplaceMe (Cosine) &  & 44.31 & 66.76 & 50.83 & \textbf{37.60} & \textbf{40.12} & 62.11 & 27.05 & 27.54 & 30.20 & \textbf{44.49} & \textbf{26.80} & \textbf{57.90} & \textbf{42.98} & \textbf{75.81\%} \\
\rowcolor{lightgreen}\methodname &  & 41.50 & 66.97 & 52.49 & 37.20 & 39.51 & \textbf{63.06} & \textbf{32.68} & 25.41 & 30.46 & 40.61 & 24.02 & 55.00 & 42.41 & 74.81\% \\

\midrule

\rowcolor{lightgrey}\texttt{\textbf{DeepSeek-7B-chat}} & Dense & 70.56 & 77.64 & 74.90 & 44.80 & 50.46 & 83.98 & 37.21 & 50.90 & 41.55 & 62.63 & 27.54 & 78.40 & 58.38 & 100.00\% \\
Streamline (FFN) &  & 25.86 & 52.45 & 63.30 & 30.40 & 36.54 & 66.12 & 24.72 & 23.05 & 27.13 & 27.02 & 17.39 & 21.40 & 34.62 & 59.29\% \\
Streamline (Layer) &  & 38.66 & 58.43 & 64.25 & \textbf{35.00} & 38.54 & 63.73 & 23.87 & 23.00 & \textbf{27.30} & 34.89 & 20.74 & 29.00 & 38.12 & 65.29\% \\
ReplaceMe (LS) & 37.5\% & 38.24 & 66.27 & 51.93 & 30.40 & 40.23 & 61.65 & \textbf{26.32} & 25.05 & 24.23 & 40.49 & 22.91 & 56.90 & 40.39 & 69.18\% \\
ReplaceMe (Cosine) &  & \textbf{42.50} & \textbf{67.79} & 52.49 & 32.00 & 39.51 & 61.47 & 25.21 & 24.54 & 25.77 & \textbf{44.11} & \textbf{23.82} & \textbf{61.10} & \textbf{41.69} & \textbf{71.41\%} \\
\rowcolor{lightgreen}\methodname &  & 37.01 & 58.38 & \textbf{65.51} & 31.80 & \textbf{40.58} & \textbf{71.62} & 26.07 & \textbf{29.75} & 26.54 & 35.06 & 22.08 & 41.50 & 40.49 & 69.36\% \\

\midrule

\rowcolor{lightgrey}\texttt{\textbf{Llama-3.1-8B-Instruct}} & Dense & 72.52 & 79.49 & 77.82 & 49.00 & 50.20 & 83.88 & 40.39 & 68.79 & 53.58 & 75.76 & 26.80 & 91.70 & 64.16 & 100.00\% \\
Streamline (FFN) &  & 37.31 & 56.64 & 66.93 & 31.60 & 36.64 & 78.87 & 28.89 & 22.96 & 28.58 & 36.41 & 19.70 & 39.50 & 40.34 & 62.87\% \\
Streamline (Layer) &  & 42.96 & 60.45 & 69.06 & \textbf{32.20} & 38.95 & 80.03 & 30.35 & 45.24 & 31.40 & \textbf{43.56} & 19.97 & 58.20 & 46.03 & 71.74\% \\
ReplaceMe (LS) & 37.5\% & 40.07 & 58.65 & 67.64 & 30.60 & 39.41 & \textbf{82.78} & 31.33 & \textbf{48.52} & 30.46 & 41.33 & 21.27 & 58.10 & 45.85 & 71.46\% \\
ReplaceMe (Cosine) &  & 42.63 & 59.36 & 65.82 & 29.20 & 39.36 & 79.05 & 31.46 & 37.10 & 31.57 & 41.08 & \textbf{22.35} & 70.50 & 45.79 & 71.37\% \\
\rowcolor{lightgreen}\methodname &  & \textbf{47.54} & \textbf{62.57} & \textbf{70.24} & 31.00 & \textbf{41.50} & 78.65 & \textbf{32.80} & 28.42 & \textbf{33.11} & 43.31 & 21.47 & \textbf{72.60} & \textbf{46.94} & \textbf{73.15\%} \\

\bottomrule
\end{tabular}
\end{adjustbox}
\caption{Downstream performance (\%) at sparsity 37.5\%. Acc. Ret. is the ratio sparse/dense aggregate accuracy.}
\end{table*}

\clearpage

\section{Limitations and Additional Analysis}
\label{sec:additional_analysis}

\subsection{Deployed parameter and compute overhead}
\label{sec:deployed-overhead}

\methodname introduces explicit bypass parameters for selected Attention and FFN submodules. \Cref{tab:params-full} reports the resulting deployed parameter overhead, with Attention and FFN compensation reported separately and the shared FFN basis counted once per model. Added parameters remain approximately $10\%$ of the removed parameters. In contrast, ReplaceMe adds no deployed parameters because its fitted transformation is folded into an existing projection.

\Cref{tab:macs-full} reports the corresponding theoretical MAC overhead for one forward pass at sequence length 2048. Attention costs include the QKV and output projections together with the QK and AV products, while FFN costs include the three FFN projections. For \methodname, the added computation is divided into Attention and FFN residual compensation and amounts to approximately $15\%$ of the removed MACs. These theoretical estimates are reported separately from offline compression time and realized inference latency.

\newpage

\begin{table}[!ht]
\label{tab:macs-full}
\centering
\begin{adjustbox}{max width=\columnwidth}
\setlength{\tabcolsep}{2pt}
\begin{tabular}{l l c c c c c}
\toprule
Sparsity & Method & Removed (GMACs) & Added (GMACs) & Add/Remove $\downarrow$ & Added attn & Added FeedForward \\
\midrule
\multicolumn{7}{c}{\texttt{\textbf{Base models}}} \\
\midrule
\multirow{5}{*}{12.5\%} & Streamline (FFN) & 1003.7 & 250.9 & 25.0\% & 0.0 & 250.9 \\
 & Streamline (Layer) & 1465.9 & 366.5 & 25.0\% & 115.5 & 250.9 \\
 & ReplaceMe (LS) & 1465.9 & 0.0 & 0.0\% & 0.0 & 0.0 \\
 & ReplaceMe (Cosine) & 1465.9 & 0.0 & 0.0\% & 0.0 & 0.0 \\
\rowcolor{lightgreen} \cellcolor{white} & \methodname & 1465.9 & 232.4 & 15.8\% & 15.1 & 217.4 \\
\cmidrule(lr){1-7}
\multirow{5}{*}{25\%} & Streamline (FFN) & 2069.0 & 250.9 & 12.3\% & 0.0 & 250.9 \\
 & Streamline (Layer) & 3019.5 & 366.5 & 12.3\% & 115.5 & 250.9 \\
 & ReplaceMe (LS) & 3019.5 & 0.0 & 0.0\% & 0.0 & 0.0 \\
 & ReplaceMe (Cosine) & 3019.5 & 0.0 & 0.0\% & 0.0 & 0.0 \\
\rowcolor{lightgreen} \cellcolor{white} &\methodname & 3019.5 & 477.0 & 15.8\% & 30.9 & 446.1 \\
\cmidrule(lr){1-7}
\multirow{5}{*}{37.5\%} & Streamline (FFN) & 3078.8 & 250.9 & 8.3\% & 0.0 & 250.9 \\
 & Streamline (Layer) & 4483.3 & 366.5 & 8.3\% & 115.5 & 250.9 \\
 & ReplaceMe (LS) & 4483.3 & 0.0 & 0.0\% & 0.0 & 0.0 \\
 & ReplaceMe (Cosine) & 4483.3 & 0.0 & 0.0\% & 0.0 & 0.0 \\
\rowcolor{lightgreen} \cellcolor{white} &\methodname & 4483.3 & 706.9 & 15.8\% & 45.8 & 661.1 \\
\midrule
\multicolumn{7}{c}{\texttt{\textbf{Instruct models}}} \\
\midrule
\multirow{5}{*}{12.5\%} & Streamline (FFN) & 1090.1 & 272.5 & 25.0\% & 0.0 & 272.5 \\
 & Streamline (Layer) & 1528.1 & 382.0 & 25.0\% & 109.5 & 272.5 \\
 & ReplaceMe (LS) & 1528.1 & 0.0 & 0.0\% & 0.0 & 0.0 \\
 & ReplaceMe (Cosine) & 1528.1 & 0.0 & 0.0\% & 0.0 & 0.0 \\
\rowcolor{lightgreen} \cellcolor{white} & \methodname & 1528.1 & 219.1 & 14.6\% & 14.6 & 204.5 \\
\cmidrule(lr){1-7}
\multirow{5}{*}{25\%} & Streamline (FFN) & 2096.4 & 272.5 & 12.9\% & 0.0 & 272.5 \\
 & Streamline (Layer) & 2956.7 & 382.0 & 12.9\% & 109.5 & 272.5 \\
 & ReplaceMe (LS) & 2956.7 & 0.0 & 0.0\% & 0.0 & 0.0 \\
 & ReplaceMe (Cosine) & 2956.7 & 0.0 & 0.0\% & 0.0 & 0.0 \\
\rowcolor{lightgreen} \cellcolor{white} & \methodname & 2956.7 & 424.5 & 14.6\% & 28.4 & 396.1 \\
\cmidrule(lr){1-7}
\multirow{5}{*}{37.5\%} & Streamline (FFN) & 3047.3 & 272.5 & 8.9\% & 0.0 & 272.5 \\
 & Streamline (Layer) & 4295.4 & 382.0 & 8.9\% & 109.5 & 272.5 \\
 & ReplaceMe (LS) & 4295.4 & 0.0 & 0.0\% & 0.0 & 0.0 \\
 & ReplaceMe (Cosine) & 4295.4 & 0.0 & 0.0\% & 0.0 & 0.0 \\
\rowcolor{lightgreen} \cellcolor{white} & \methodname & 4295.4 & 615.2 & 14.6\% & 41.3 & 573.9 \\
\bottomrule
\end{tabular}
\end{adjustbox}
\caption{Theoretical MACs removed and added for one forward pass with sequence length 2048. Attention MACs include QKV/O projections plus QK and AV products; FeedForward MACs include the three FeedForward projections. For \methodname, added MACs are split into Attention and FeedForward residual compensation.}
\end{table}

\begin{table*}[t]
\centering
\tiny
\begin{adjustbox}{max width=\textwidth}
\setlength{\tabcolsep}{3pt}
\begin{tabular}{l l l c c c c c}
\toprule
Model & Sparsity & Method & Removed (M) & Added (M) & Add/Remove $\downarrow$ & Added Attn. (M) & Added FFN (M) \\
\midrule
\multirow{15}{*}{\texttt{\textbf{Llama-3.2-3B}}} & \multirow{5}{*}{12.5\%} & Streamline (FFN) & 402.7 & 75.5 & 18.7\% & 0.0 & 75.5 \\
 &  & Streamline (Layer) & 402.7 & 100.7 & 25.0\% & 25.2 & 75.5 \\
 &  & ReplaceMe (LS) & 402.7 & 0.0 & 0.0\% & 0.0 & 0.0 \\
 &  & ReplaceMe (Cosine) & 402.7 & 0.0 & 0.0\% & 0.0 & 0.0 \\
\rowcolor{lightgreen} \cellcolor{white}  & \cellcolor{white}  & \methodname & 402.7 & 53.6 & 13.3\% & 6.3 & 47.2 \\
\cmidrule(lr){2-8}
 & \multirow{5}{*}{25\%} & Streamline (FFN) & 704.7 & 75.5 & 10.7\% & 0.0 & 75.5 \\
 &  & Streamline (Layer) & 704.7 & 100.7 & 14.3\% & 25.2 & 75.5 \\
 &  & ReplaceMe (LS) & 704.7 & 0.0 & 0.0\% & 0.0 & 0.0 \\
 &  & ReplaceMe (Cosine) & 704.7 & 0.0 & 0.0\% & 0.0 & 0.0 \\
\rowcolor{lightgreen} \cellcolor{white}  & \cellcolor{white}  & \methodname  & 704.7 & 86.6 & 12.3\% & 11.1 & 75.6 \\
\cmidrule(lr){2-8}
 & \multirow{5}{*}{37.5\%} & Streamline (FFN) & 1006.7 & 75.5 & 7.5\% & 0.0 & 75.5 \\
 &  & Streamline (Layer) & 1006.7 & 100.7 & 10.0\% & 25.2 & 75.5 \\
 &  & ReplaceMe (LS) & 1006.7 & 0.0 & 0.0\% & 0.0 & 0.0 \\
 &  & ReplaceMe (Cosine) & 1006.7 & 0.0 & 0.0\% & 0.0 & 0.0 \\
\rowcolor{lightgreen} \cellcolor{white}  & \cellcolor{white}  & \methodname & 1006.7 & 119.7 & 11.9\% & 15.8 & 103.9 \\
\midrule
\multirow{15}{*}{\texttt{\textbf{Qwen3-4B}}} & \multirow{5}{*}{12.5\%} & Streamline (FFN) & 403.7 & 74.7 & 18.5\% & 0.0 & 74.7 \\
 &  & Streamline (Layer) & 403.7 & 100.9 & 25.0\% & 26.2 & 74.7 \\
 &  & ReplaceMe (LS) & 403.7 & 0.0 & 0.0\% & 0.0 & 0.0 \\
 &  & ReplaceMe (Cosine) & 403.7 & 0.0 & 0.0\% & 0.0 & 0.0 \\
\rowcolor{lightgreen} \cellcolor{white}  & \cellcolor{white}  & \methodname & 403.7 & 38.1 & 9.4\% & 5.3 & 32.8 \\
\cmidrule(lr){2-8}
 & \multirow{5}{*}{25\%} & Streamline (FFN) & 908.4 & 74.7 & 8.2\% & 0.0 & 74.7 \\
 &  & Streamline (Layer) & 908.4 & 100.9 & 11.1\% & 26.2 & 74.7 \\
 &  & ReplaceMe (LS) & 908.4 & 0.0 & 0.0\% & 0.0 & 0.0 \\
 &  & ReplaceMe (Cosine) & 908.4 & 0.0 & 0.0\% & 0.0 & 0.0 \\
\rowcolor{lightgreen} \cellcolor{white}  & \cellcolor{white}  & \methodname  & 908.4 & 77.5 & 8.5\% & 11.9 & 65.6 \\
\cmidrule(lr){2-8}
 & \multirow{5}{*}{37.5\%} & Streamline (FFN) & 1413.0 & 74.7 & 5.3\% & 0.0 & 74.7 \\
 &  & Streamline (Layer) & 1413.0 & 100.9 & 7.1\% & 26.2 & 74.7 \\
 &  & ReplaceMe (LS) & 1413.0 & 0.0 & 0.0\% & 0.0 & 0.0 \\
 &  & ReplaceMe (Cosine) & 1413.0 & 0.0 & 0.0\% & 0.0 & 0.0 \\
\rowcolor{lightgreen} \cellcolor{white}  & \cellcolor{white}  & \methodname  & 1413.0 & 116.9 & 8.3\% & 18.5 & 98.4 \\
\midrule
\multirow{15}{*}{\texttt{\textbf{DeepSeek-7B}}} & 12.5\% & Streamline (FFN) & 809.5 & 135.3 & 16.7\% & 0.0 & 135.3 \\
 &  & Streamline (Layer) & 809.5 & 202.4 & 25.0\% & 67.1 & 135.3 \\
 &  & ReplaceMe (LS) & 809.5 & 0.0 & 0.0\% & 0.0 & 0.0 \\
 &  & ReplaceMe (Cosine) & 809.5 & 0.0 & 0.0\% & 0.0 & 0.0 \\
\rowcolor{lightgreen} \cellcolor{white}  & \cellcolor{white}  & \methodname  & 809.5 & 92.4 & 11.4\% & 8.4 & 83.9 \\
\cmidrule(lr){2-8}
 & \multirow{5}{*}{25\%} & Streamline (FFN) & 1619.1 & 135.3 & 8.4\% & 0.0 & 135.3 \\
 &  & Streamline (Layer) & 1619.1 & 202.4 & 12.5\% & 67.1 & 135.3 \\
 &  & ReplaceMe (LS) & 1619.1 & 0.0 & 0.0\% & 0.0 & 0.0 \\
 &  & ReplaceMe (Cosine) & 1619.1 & 0.0 & 0.0\% & 0.0 & 0.0 \\
\rowcolor{lightgreen} \cellcolor{white}  & \cellcolor{white}  & \methodname  & 1619.1 & 168.0 & 10.4\% & 16.9 & 151.1 \\
\cmidrule(lr){2-8}
 & \multirow{5}{*}{37.5\%} & Streamline (FFN) & 2226.2 & 135.3 & 6.1\% & 0.0 & 135.3 \\
 &  & Streamline (Layer) & 2226.2 & 202.4 & 9.1\% & 67.1 & 135.3 \\
 &  & ReplaceMe (LS) & 2226.2 & 0.0 & 0.0\% & 0.0 & 0.0 \\
 &  & ReplaceMe (Cosine) & 2226.2 & 0.0 & 0.0\% & 0.0 & 0.0 \\
\rowcolor{lightgreen} \cellcolor{white}  & \cellcolor{white}  & \methodname  & 2226.2 & 224.7 & 10.1\% & 23.2 & 201.5 \\
\midrule
\multirow{15}{*}{\texttt{\textbf{Llama-3.1-8B}}} & \multirow{5}{*}{12.5\%} & Streamline (FFN) & 872.4 & 176.2 & 20.2\% & 0.0 & 176.2 \\
 &  & Streamline (Layer) & 872.4 & 218.1 & 25.0\% & 41.9 & 176.2 \\
 &  & ReplaceMe (LS) & 872.4 & 0.0 & 0.0\% & 0.0 & 0.0 \\
 &  & ReplaceMe (Cosine) & 872.4 & 0.0 & 0.0\% & 0.0 & 0.0 \\
\rowcolor{lightgreen} \cellcolor{white}  & \cellcolor{white}  & \methodname  & 872.4 & 92.4 & 10.6\% & 8.4 & 83.9 \\
\cmidrule(lr){2-8}
 & \multirow{5}{*}{25\%} & Streamline (FFN) & 1744.9 & 176.2 & 10.1\% & 0.0 & 176.2 \\
 &  & Streamline (Layer) & 1744.9 & 218.1 & 12.5\% & 41.9 & 176.2 \\
 &  & ReplaceMe (LS) & 1744.9 & 0.0 & 0.0\% & 0.0 & 0.0 \\
 &  & ReplaceMe (Cosine) & 1744.9 & 0.0 & 0.0\% & 0.0 & 0.0 \\
\rowcolor{lightgreen} \cellcolor{white}  & \cellcolor{white}  & \methodname  & 1744.9 & 168.0 & 9.6\% & 16.9 & 151.1 \\
\cmidrule(lr){2-8}
 & \multirow{5}{*}{37.5\%} & Streamline (FFN) & 2617.3 & 176.2 & 6.7\% & 0.0 & 176.2 \\
 &  & Streamline (Layer) & 2617.3 & 218.1 & 8.3\% & 41.9 & 176.2 \\
 &  & ReplaceMe (LS) & 2617.3 & 0.0 & 0.0\% & 0.0 & 0.0 \\
 &  & ReplaceMe (Cosine) & 2617.3 & 0.0 & 0.0\% & 0.0 & 0.0 \\
\rowcolor{lightgreen} \cellcolor{white}  & \cellcolor{white}  & \methodname  & 2617.3 & 243.6 & 9.3\% & 25.3 & 218.3 \\
\midrule
\multirow{15}{*}{\texttt{\textbf{Qwen3-8B}}} & \multirow{5}{*}{12.5\%} & Streamline (FFN) & 771.8 & 151.0 & 19.6\% & 0.0 & 151.0 \\
 &  & Streamline (Layer) & 771.8 & 192.9 & 25.0\% & 41.9 & 151.0 \\
 &  & ReplaceMe (LS) & 771.8 & 0.0 & 0.0\% & 0.0 & 0.0 \\
 &  & ReplaceMe (Cosine) & 771.8 & 0.0 & 0.0\% & 0.0 & 0.0 \\
\rowcolor{lightgreen} \cellcolor{white}  & \cellcolor{white}  & \methodname  & 771.8 & 92.4 & 12.0\% & 8.4 & 83.9 \\
\cmidrule(lr){2-8}
 & \multirow{5}{*}{25\%} & Streamline (FFN) & 1736.5 & 151.0 & 8.7\% & 0.0 & 151.0 \\
 &  & Streamline (Layer) & 1736.5 & 192.9 & 11.1\% & 41.9 & 151.0 \\
 &  & ReplaceMe (LS) & 1736.5 & 0.0 & 0.0\% & 0.0 & 0.0 \\
 &  & ReplaceMe (Cosine) & 1736.5 & 0.0 & 0.0\% & 0.0 & 0.0 \\
\rowcolor{lightgreen} \cellcolor{white}  & \cellcolor{white}  & \methodname  & 1736.5 & 186.9 & 10.8\% & 19.0 & 167.9 \\
\cmidrule(lr){2-8}
 & \multirow{5}{*}{37.5\%} & Streamline (FFN) & 2701.2 & 151.0 & 5.6\% & 0.0 & 151.0 \\
 &  & Streamline (Layer) & 2701.2 & 192.9 & 7.1\% & 41.9 & 151.0 \\
 &  & ReplaceMe (LS) & 2701.2 & 0.0 & 0.0\% & 0.0 & 0.0 \\
 &  & ReplaceMe (Cosine) & 2701.2 & 0.0 & 0.0\% & 0.0 & 0.0 \\
\rowcolor{lightgreen} \cellcolor{white}  & \cellcolor{white}  & \methodname  & 2701.2 & 281.4 & 10.4\% & 29.5 & 251.8 \\
\bottomrule
\end{tabular}
\end{adjustbox}
\caption{Removed parameters, deployed added compensation parameters, and added/removed ratio for base models. For \methodname, the table reports the Attention and FFN compensation breakdown; the FFN replacement uses a shared basis counted once per model. Streamline FFN removes the same decoder-layer budget as the other depth-replacement methods but deploys an FFN-only replacement. ReplaceMe folds its fitted full-rank transform into an existing FeedForward projection, so it adds no deployed parameters.}
\label{tab:params-full}
\end{table*}

\stopcontents[appendices]
\end{document}